\documentclass{article}

    \PassOptionsToPackage{numbers, compress}{natbib}

\usepackage[final]{neurips_2023}

\usepackage[utf8]{inputenc} 
\usepackage[T1]{fontenc}    
\usepackage{hyperref}       
\usepackage{url}            
\usepackage{booktabs}       
\usepackage{amsfonts}       
\usepackage{nicefrac}       
\usepackage{microtype}      
\usepackage{xcolor}         

\usepackage{enumitem}
\usepackage{xspace}
\usepackage{bm}
\usepackage{hyperref}
\usepackage{url}
\usepackage{graphicx}
\usepackage{amsmath,amssymb,amsfonts}
\usepackage[ruled,vlined]{algorithm2e}
\usepackage{color}
\usepackage{amsthm}
\usepackage{hyperref}
\usepackage{subfigure}
\usepackage{mathrsfs}
\usepackage{enumitem}
\usepackage{bbm}
\usepackage{wrapfig}
\usepackage{makecell}
\usepackage{multirow}
\usepackage{colortbl}
\usepackage{booktabs}
\usepackage{pifont}
\usepackage{colortbl}
\usepackage{graphicx}

\definecolor{mygraylite}{gray}{.94}
\definecolor{mygray}{gray}{.89}
\definecolor{darkgreen}{rgb}{0, 0.5, 0}
\definecolor{darkergreen}{RGB}{21, 152, 56}
\definecolor{amber}{rgb}{1.0, 0.75, 0.0}
\definecolor{darkseagreen}{rgb}{0.56, 0.74, 0.56}
\definecolor{darkblue}{rgb}{0, 0, 54.5}
\definecolor{darkorange}{rgb}{220,88,42}
\usepackage{makecell}

\newcommand*{\modelname}{\text{TPSR}\@\xspace}

\newlist{myitemize2}{itemize}{4}
\setlist[myitemize2,1]{label={},leftmargin=0em}
\newlist{myitemize}{itemize}{4}
\setlist[myitemize,1]{label=\textbullet,leftmargin=1em}

\title{Transformer-based Planning for Symbolic Regression}

\author{
\textbf{Parshin Shojaee}\thanks{Equal contribution. Contact email: parshinshojaee@vt.edu} {}$^{ \ 1}$
,  Kazem Meidani$^{* \ 2}$\\
\textbf{Amir Barati Farimani}$^{ \ 2,3}$
\textbf{, Chandan K. Reddy}$^{1}$\\
$^{1}$ Department of Computer Science, Virginia Tech\\
$^{2}$  Department of Mechanical Engineering, Carnegie Mellon University\\
$^{3}$  Machine Learning Department, Carnegie Mellon University\\
}

\begin{document}

\maketitle

\begin{abstract}
Symbolic regression (SR) is a challenging task in machine learning that involves finding a mathematical expression for a function based on its values. Recent advancements in SR have demonstrated the effectiveness of pre-trained transformer-based models in generating equations as sequences, leveraging large-scale pre-training on synthetic datasets and offering notable advantages in terms of inference time over classical Genetic Programming (GP) methods. However, these models primarily rely on supervised pre-training goals borrowed from text generation and overlook equation discovery objectives like accuracy and complexity. To address this, we propose \modelname, a \textbf{T}ransformer-based \textbf{P}lanning strategy for \textbf{S}ymbolic \textbf{R}egression that incorporates Monte Carlo Tree Search into the transformer decoding process. Unlike conventional decoding strategies, \modelname enables the integration of non-differentiable feedback, such as fitting accuracy and complexity, as external sources of knowledge into the transformer-based equation generation process. Extensive experiments on various datasets show that our approach outperforms state-of-the-art methods, enhancing the model's fitting-complexity trade-off, extrapolation abilities, and robustness to noise
\footnote{The codes are available at: \url{https://github.com/deep-symbolic-mathematics/TPSR}}
. 
\end{abstract}

\vspace{-0.5em}
\section{Introduction}
\label{sec:intro}
\vspace{-0.5em}
Symbolic regression (SR) is a powerful method to discover mathematical expressions for governing equations of complex systems and to describe data patterns in an interpretable symbolic form. It finds extensive applications in science and engineering, enabling the modeling of physical phenomena in various domains such as molecular dynamics, fluid dynamics, and cosmology \cite{MD-Functional-Science-2022, materials-sym-2019, Rudy-PDE-Science-2017,Meidani-PDE-2021,SINDY-PNAS-2016, Cranmer-cosmology-NeurIPS-2020}. 
Symbolic representations provide valuable insights into complex systems, facilitating a better understanding, prediction, and control of these systems  through the design of accurate, generalizable, and efficient models \citep{Sindy-MPC-2018, Meidani-IP2-2023, rethinking-scientific-2022}. 
SR models establish the functional relationship between independent and target variables by mapping them to mathematical equations. The input data can be obtained from simulations, experimental measurements, or real-world observations. Symbolic regression, however, poses several challenges, including the combinatorial nature of the optimization search space, vulnerability to the quality of input data, and the difficulty of striking a balance between model fitting, complexity, and generalization performance \citep{NP-hard-Symbolic-2022, Console-convex-NIPS-2022}. 


Symbolic regression encompasses a wide range of methods, spanning different categories. Traditional approaches, such as Genetic Programming (GP), use a heuristic population-based search strategy where each individual represents a potential solution to the problem \citep{Schmidt-Lipson-2009, operon-GP-2020}. Though GP algorithms are capable of finding solutions for nonlinear and complex problems, they are typically slow to converge 
due to the vast functional search space. Also, as they need to start the search from scratch for each dataset, they tend to be computationally expensive, prone to overfitting, and sensitive to the choice of parameters \citep{DSR-Petersen-ICLR-2021}. Recent works in SR have shown promising results by using pre-trained transformers \citep{Attention-NeurIPS-2017} for generating equations as sequences of tokens. These models leverage the prior knowledge learned through large-scale pre-training and can generate equations with a single forward pass, leading to faster inference times compared to GP-based methods \citep{Biggio-NeSymReS-ICML-2021, SymbolicGPT-arXiv-2021, Kamienny-E2E-symbolic-NIPS-2022,symformer-arxiv-2022}. 
However, one of the limitations of these models is that they focus on the supervised pre-training goals borrowed from text generation, i.e., they are trained solely with the token-level cross-entropy (CE) loss, which can result in equations that may exhibit high token-level similarities but are suboptimal with respect to equation discovery objectives such as fitting accuracy and complexity. To mitigate this issue, beam search \citep{beam-search-2012, beam-search-2016} or sampling \citep{sampling-2018-hierarchical} approaches are employed as decoding strategies to propose multiple candidate equations for a given dataset, and then select the optimal candidate equation based on the fitting accuracy after optimizing for constants. Nonetheless, both beam search and sampling decoding strategies primarily rely on the pre-trained transformer's logits and next token probability distributions, and therefore do not receive any performance feedback during the generation of equation candidates.


\begin{figure}[t]
\centering
\includegraphics[width=\linewidth] 
{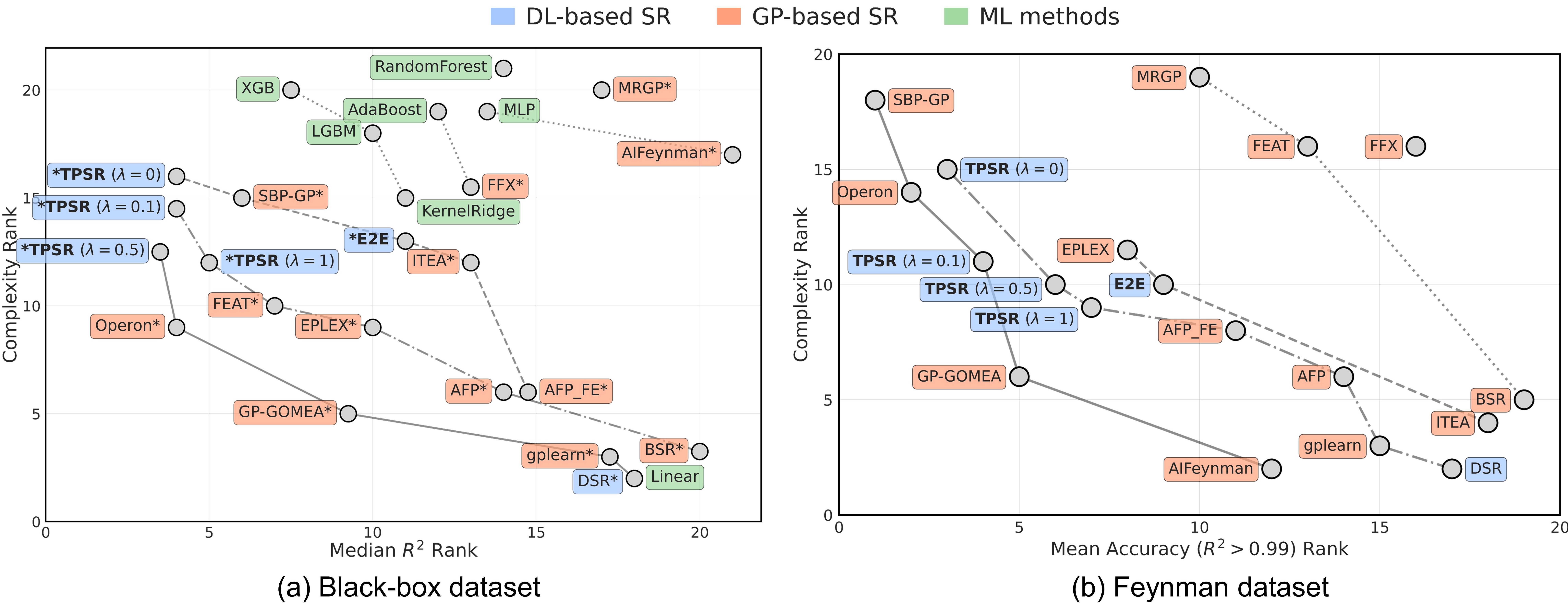}
\caption{Pareto plot comparing the rankings of
all methods in terms of the $R^2$ performance and identified equation complexity for \textbf{(a) SRBench \textit{Black-box} datasets} and \textbf{(b) \textit{Feynman} datasets}. Our results with Transformer-based Planning (TPSR) applied on top of E2E transformer SR model improves its average accuracy on both data groups while maintaining a similar range of equation complexity. \textit{\modelname can successfully reach the first Pareto-front which is better than E2E baseline on both data groups}.
Connecting lines denote Pareto dominance
rankings, colors denote the families of models, and "$*$" indicates SR methods in \textit{Black-box} datasets.
\vspace{-0.5em}
}
\label{fig:pareto-plots}
\end{figure}


To consider the equation discovery objectives in the transformer generation process and still benefit from the pre-trained model logits, we propose \modelname, a \textbf{T}ransformer-based \textbf{P}lanning strategy for \textbf{S}ymbolic \textbf{R}egression. \modelname leverages a lookahead planning algorithm, using Monte Carlo Tree Search (MCTS) as a decoding strategy on top of pre-trained transformer-based SR models to guide equation sequence generation.
\modelname significantly improves the performance of generated equations by considering feedback during the generation process and still remains faster than GP-based models which do not leverage the pre-training priors and learn expressions for each dataset from scratch. 
Notably, our approach is model-agnostic and can be applied to any pre-trained SR model, enabling optimization of generated equation sequences for non-differentiable objectives that may encompass combinations of fitting accuracy, complexity, and equation forms. Additionally, we incorporate different caching mechanisms to reduce the overall inference time. Our experimental results demonstrate that applying \modelname on top of the pre-trained E2E SR model \citep{Kamienny-E2E-symbolic-NIPS-2022} significantly enhances its performance across various benchmark datasets. 
As depicted in Fig. \ref{fig:pareto-plots}, \modelname achieves a strong balance between fitting accuracy and model complexity compared to other leading baselines. It also effectively drives the E2E model towards the optimal trade-off, represented by the first Pareto front. The major contributions of this work are summarized below:

\vspace{-0.5em}
\begin{itemize}[leftmargin=*]
\item Proposing \modelname, a new method that combines pre-trained transformer SR models with Monte Carlo Tree Search (MCTS) lookahead planning to optimize the generation of equation sequences while considering non-differentiable performance feedback.
\item Developing a new reward function that balances equation fitting accuracy and complexity to optimize the generated equations for an effective trade-off.
\item Demonstrating that \modelname consistently outperforms state-of-the-art baselines across various SR benchmark datasets, generating equations with higher fitting accuracy while maintaining lower complexity to avoid non-parsimonious solutions.
\item Showcasing the extrapolation and noise robustness of \modelname compared to the baseline and conducting an ablation study to investigate the impact of various model components.
\end{itemize}


\vspace{-0.5em}
\section{Related Work}
\label{sec:related}
\vspace{-0.5em}
\paragraph{Symbolic Regression without Learned Priors.}
Genetic Programming (GP) algorithms are typically employed for single-instance SR, aiming to find the best-fit equation for a "single" dataset at hand \citep{Schmidt-Lipson-2009}. 
Recently, alternative neural network-based search algorithms have been explored, including deep reinforcement learning (RL) \citep{DSR-Petersen-ICLR-2021,rl_sr2,symbolic-RL-ICML-2021}, combinations of GP and RL \citep{mundhenk-seeding-GP-NeurIPS-2021}, and Monte Carlo Tree Search (MCTS) as a standalone framework \citep{SPL-MCTS-ICLR-2023}. Despite their successes, all these methods lack the benefits of prior knowledge learned from large-scale pre-training. Consequently, they are slow during inference as they need to restart the search from scratch for new datasets.

\vspace{-0.2em}
\paragraph{Pre-trained Transformers for Symbolic Regression.}
In recent years, pre-trained transformers have shown remarkable performance in natural language and programming language tasks \cite{devlin-etal-2019-bert, gpt_neurips2016, wang-etal-2021-codet5}. This success has inspired researchers to develop pre-trained transformer models for SR \citep{Biggio-NeSymReS-ICML-2021, SymbolicGPT-arXiv-2021, Kamienny-E2E-symbolic-NIPS-2022, symformer-arxiv-2022,landajuela2022a}. For example, Biggio \textit{et al.} \citep{Biggio-NeSymReS-ICML-2021} introduced a Neural Symbolic Regression model that scales (NeSymReS) with the amount of synthetic training data and generates equation skeletons where all the numerical constants are represented by a single token ``$C$''. 
Kamienny \textit{et al.} \citep{Kamienny-E2E-symbolic-NIPS-2022} proposed an end-to-end framework that predicts the complete equation form along with its constants.
More recent works \citep{landajuela2022a, holt2023dgsr} introduced unified frameworks that include a transformer-based pre-training stage as the prior for subsequent RL or GP optimization steps.
While GP and RL methods have to start anew for each problem, the purely transformer-based approaches rely on synthetic data and the power of large-scale pre-trained priors to generate equations in a single forward pass. However, these models are mostly pre-trained on token-level sequence generation losses, and thus can perform suboptimal for other equation discovery objectives such as fitting accuracy and complexity. 
Our model, \modelname, utilizes lookahead planning to guide the generation of equations towards better performance by employing fitting and complexity feedback during the transformer generation process.

\vspace{-0.2em}
\paragraph{Planning in Sequence Generation.}
Recently, planning algorithms 
have been utilized in NLP tasks to optimize text output for specific objectives, such as controlling generated text to meet certain constraints like non-toxicity or conveying certain emotions \citep{CoopMCTS_NLP_NeurIPS2021,leblond-acl2021,chaffin-acl2022}.
Recent advances in programming language models developed in code generation have also yielded promising techniques that could be adapted for SR, as they share several vital similarities with each other. Both involve generating sequences of symbols for a given input and typically require optimizing the generated sequences for specific criteria which is different from the pre-trianing objective. For code generation, this may involve optimizing objectives like code compilability, readability, or passing test cases \citep{chen2018executionguided,rl_code1,MCTS-code-ICLR-2023}. Similarly, in SR, the focus may be on equation-specific sequence-level objectives such as fitting accuracy or minimizing complexity. Motivated by these successes, we develop an approach that combines 
MCTS
planning with pre-trained transformer SR models for improved equation discovery.

\vspace{-0.5em}
\section{Methodology}
\label{sec:method}
\vspace{-0.5em}
\subsection{Preliminaries}
\begin{figure}[t]
\centering
\includegraphics[width=0.99\linewidth]{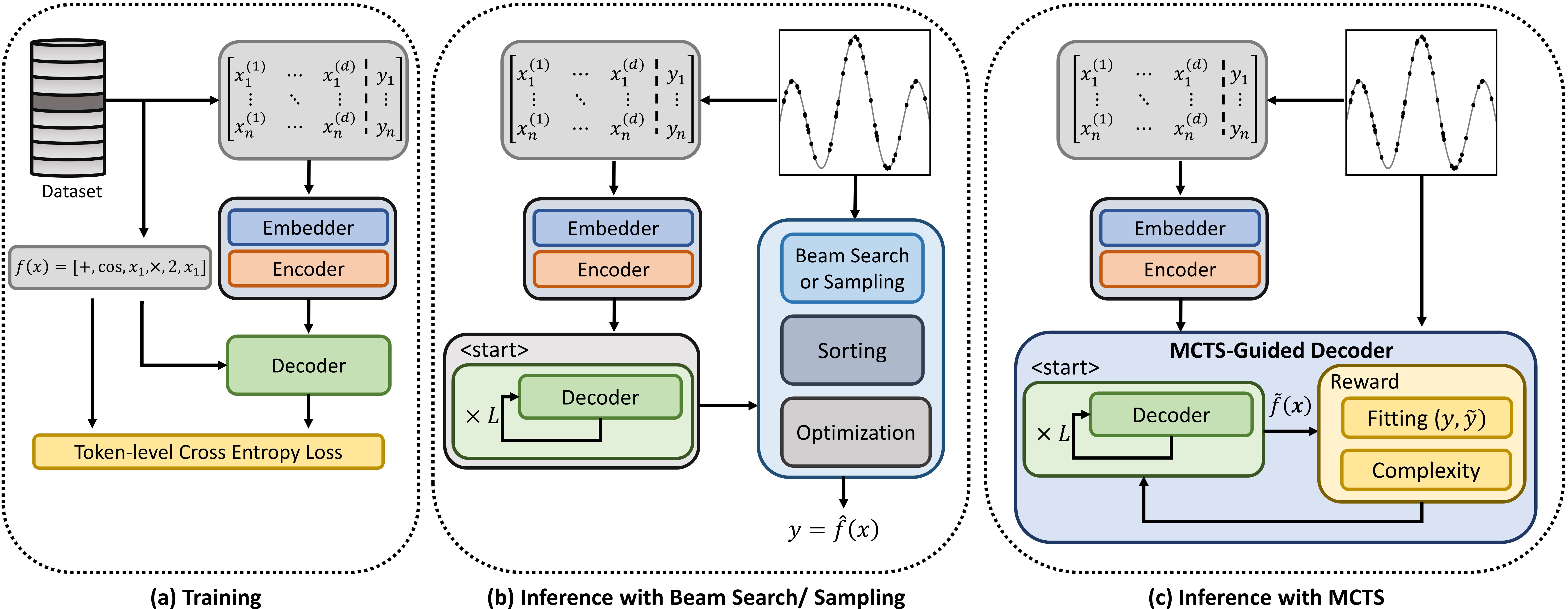}
\caption{An overview of our proposed method with MCTS-guided decoding at inference compared to the concurrent works with beam search/sampling decoding strategy.
}
\label{fig:deepsr}
\end{figure}

\vspace{-0.5em}
In SR, the main goal is to find a symbolic expression for the unknown function $f(\cdot)$ mapping the $d$-dimensional input $\bm x \in \mathbb{R}^{d}$ to the target variable $ y = f(\bm x) \in \mathbb{R} $. Given a dataset of $n$ observations $\mathcal{D} = (\bm x_i, y_i)_{i=1}^{n}$, SR methods try to generate an equation $\tilde{f}(\cdot)$ such that $y_i \approx \tilde{f}(\bm x_i)$ for all $i \in \mathbb{N}_n$. Also, the proposed equation is desired to generalize well and to effectively balance the fitting accuracy and complexity.
The transformer-based SR models are trained on a large-scale dataset comprising equation instances paired with their corresponding observations,
$\{ (\mathcal{D}_1, f_1(\cdot)) \; \ldots$ $\; (\mathcal{D}_M, f_M(\cdot)) \}$, where $M$ is the dataset size. 
During inference, the trained model directly generates the equation $\tilde{f}(\cdot)$ as a sequence of tokens in an autoregressive manner.
An effective way to represent the expression tree of equations in a sequence is to use prefix notation as in \citep{Lample-Deep-SR-ICLR-2020}.
For embedding the observations, we adopt the pre-trained SR model backbone from \cite{Kamienny-E2E-symbolic-NIPS-2022}. Notably, given the potential for large input sequences with tokenized numeric data and the quadratic complexity of Transformers, the method introduced in \cite{Kamienny-E2E-symbolic-NIPS-2022} deploys a linear embedder module to map inputs to a singular embedding space before introducing them to the Transformers encoder. Subsequent to embedding, transformer-based SR models encode the input observations and then pass the encoded representation along with the masked tokens to decode the equation sequence. To train the model, token-level cross-entropy loss is employed to learn the distribution of next token prediction conditioned on the encoded dataset and the current state of sequence (Fig.~\ref{fig:deepsr}(a)).

Achieving a good fitting performance from the model's predicted sequence demands generating accurate constants in the equation. To address this, the generated skeleton or equation can undergo a round of optimization to estimate their constants using nonlinear methods, such as Broyden–Fletcher–Goldfarb–Shanno algorithm (BFGS) \citep{bfgs_flet87}. 
Previous works \citep{Kamienny-E2E-symbolic-NIPS-2022, Biggio-NeSymReS-ICML-2021} employ beam search and sampling strategies for transformer decoding in combination with constant optimization to propose several candidate equations. Subsequently, they use fitting metrics such as $R^2$ to order these candidates and output the final equation with the best performance (Fig.~\ref{fig:deepsr}(b)). Transformer models utilizing beam search or sampling decoding strategies can generate multiple high-likelihood equation sequences, but they rely on logits obtained from model parameters pre-trained with token-matching loss relative to the reference equation. As a result, such models lack the capability to receive feedback and optimize generation for equation discovery objectives such as fitting or complexity of equations.

\vspace{-0.5em}
\subsection{MCTS-Guided Equation Generation}
\vspace{-0.5em}
To generate equations that are both better-fitting and less-complex, it is crucial to incorporate feedback into the equation generation process. To achieve this, we utilize Monte Carlo Tree Search (MCTS) lookahead planning during inference, guiding the decoder towards optimal solutions for fitting and complexity objectives (as shown in Fig. \ref{fig:deepsr}(c)). 
The MCTS-guided transformer decoding explores different possibilities, identifying the most promising paths based on the objectives.

We frame the SR equation generation task as a Markov Decision Process (MDP) where state $s$ represents the current sequence at generation step (token) $t$. If $s$ has not reached the terminal state (i.e., the <\texttt{EOS}> token), we select the next token from the vocabulary as action $a$, updating state $s'$ by concatenating $s$ and $a$. Upon reaching the terminal state, the reward $r$ is computed and used to update the decoding model. MCTS represents states as nodes and actions as edges within a tree structure, navigating state-space from the root node (i.e., initial state) to reach terminal states with maximum rewards.
MCTS balances exploration and exploitation, considering nodes with higher quality equations (i.e., higher Q-values) and under-explored nodes (i.e., those with fewer visits).
During the generation process of the transformer, we utilize the MCTS algorithm iteratively to conduct lookahead planning and determine the next token. However, the large search-space requires more than the sole application of MCTS to generate high-quality equations.  We need to effectively share information between the pre-trained transformer model and MCTS for better generations. To achieve this, we incorporate the probabilities of the next-token that are acquired through the pre-trained transformer SR models into the MCTS planning process. This incorporation helps to enhance the search process, leading to more efficient and effective results. 
The key steps of MCTS for transformer decoding in SR models, as depicted in Fig. \ref{fig:mcts}, are as follows:

\begin{figure}[t]
\centering
\includegraphics[width=0.9\linewidth]{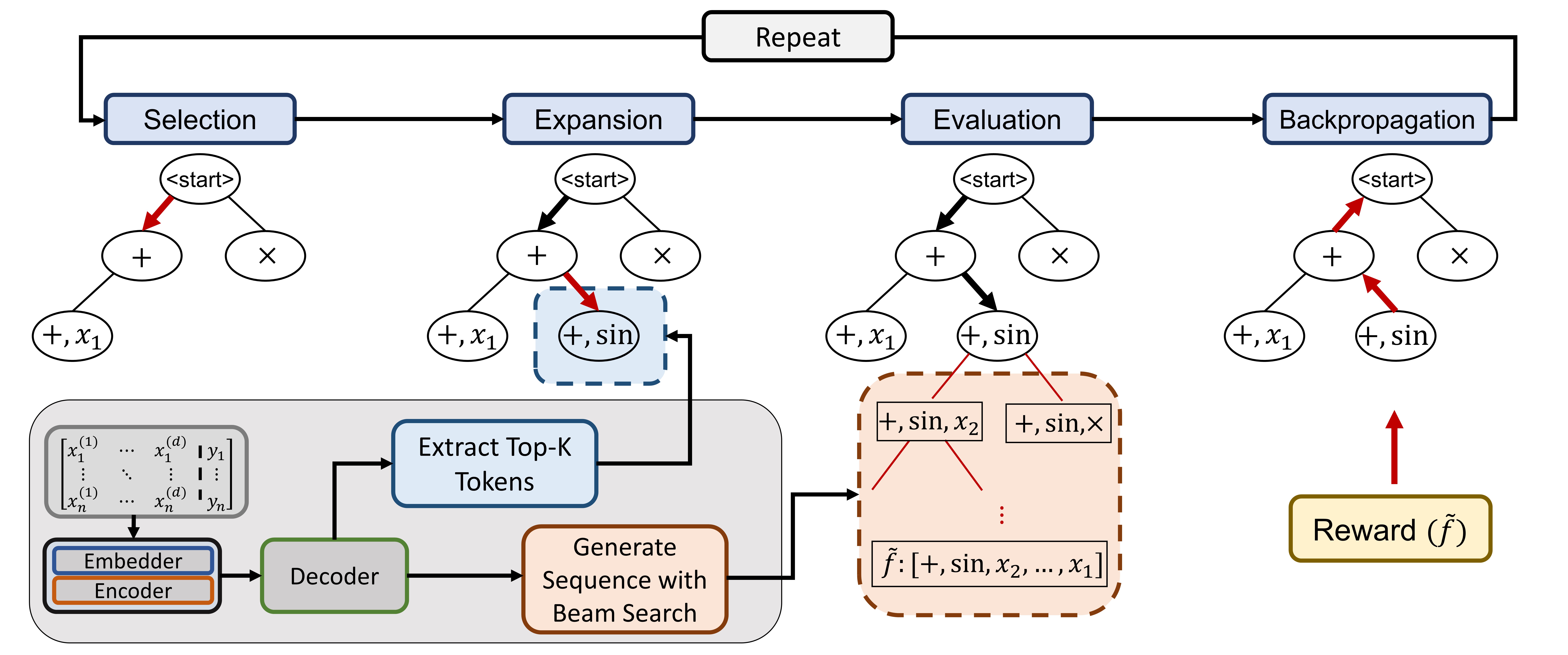}
\caption{Overview of \modelname's key steps: Selection, Expansion, Evaluation, and Backpropagation. MCTS-guided decoding interacts with the pre-trained transformer SR model in the expansion and evaluation steps employing the transformer $top$-$k$ sampling and beam search, respectively. The designed reward is used to guide the backpropagation. 
}
\label{fig:mcts}
\end{figure}


\begin{myitemize2}
\item
{\textbf{Selection.}} 
The Upper Confidence Bound for Trees \citep{UCB} criterion is employed to select actions (i.e., next tokens) for fully extended nodes in the search tree, balancing exploration and exploitation. We use the P-UCB heuristic in \cite{silveralphazero} as

\vspace{-1.0em}
\begin{equation}
\mathrm{P\text{-}UCB}(s,a) = Q(s,a) + \beta \cdot P_{\theta}(a|s) \cdot \sqrt{\frac{\ln\left(N(s)\right)}{1+N(s')}},
\label{eq:uct}
\end{equation}

where $Q(s,a)$ is the maximum return for action $a$ in state $s$ across all simulations, promoting the exploitation of the optimal child node. The second term encourages exploration of less-visited children, with $N(s)$ as state $s$'s visit count and $s'$ as the subsequent state. $P_{\theta}(a|s)$ is the probability of the next token $a$ given the partial sequence state $s$ from pre-trained transformer model parameterized by $\theta$. The exploration-exploitation trade-off is adjusted by hyperparameter $\beta$.
Lastly, the next token action maximizes the P-UCB: $\mathrm{Select}(s) = \arg{\max_{a}{\mathrm{P\text{-}UCB}(s,a)}}$. 


\item \textbf{Expansion.} 
In the expansion stage, after selecting a node that is not fully expanded, a new child (next token) for the current state is explored. Random expansion of the node from the vocabulary, however, might result in an invalid equation (that does not comply with the prefix notation) and makes the search process very time-consuming. Therefore, given partial equations, only $top$-$k$ most likely choices of the next token are considered as the possible children of the node for expansion. In other words, we are restricting the actions to be only from the $top$-$k$ high-likelihood options which are retrieved from the pre-trained transformer SR model's logits. These options are then ordered to determine the sequence in which the children will be expanded. 

\item {\textbf{Evaluation.}} To evaluate the newly expanded nodes, we perform simulations to complete the equation sequence. This is necessary because the new state may still be a partial equation and performance feedback can only be obtained at the end of the sequence when the equation generation is completed. 
In MCTS, it is common to employ random actions during the simulation stage. Nevertheless, random action selection for equation generation, much like during expansion, suffers from certain drawbacks in terms of time and the possibility of generating invalid equations. Consequently, the pre-trained transformer SR model is invoked again, this time utilizing beam search with a beam size of $b$, to generate complete equation candidates based on the current state. The beam size $b$ determines the number of complete equations to be generated from the current partial equation. Following the simulations, the highest reward among all the candidates is assigned to the new node value. 

\item {\textbf{Backpropagation.}} After generating a complete equation $\tilde{f}(\cdot)$, the corresponding reward $r(\tilde{f}(\cdot))$ can be computed. The highest reward among all simulations is then assigned to the new node, which recursively backpropagates its estimated value to its parents until it reaches the root of the tree. This update process involves updating the $Q$ values of all state-action pairs, denoted as $s'$ and $a'$, along the trajectory in the tree to reach the root. Specifically, for each state-action pair, the $Q$ value is updated by taking the maximum of the current $Q$ value and the new value $r$: $Q(s',a') \leftarrow \max{(Q(s',a'),r)}$.

\vspace{0.1em}
More details on \modelname, including its steps and implementation can be found in Appendix \ref{sec:app_methods}.

\end{myitemize2}

\vspace{-0.5em}
\subsection{Reward Definition}
\vspace{-0.5em}
We define a numerical reward $r \in \mathbb{R}$ to evaluate complete equation candidate $\tilde{f}(\cdot)$, promoting fitting accuracy and regulating complexity. After optimizing constants in the complete sequence, we compute the reward. We first calculate the normalized mean squared error (NMSE) between ground-truth target variable $y$ and predicted target variable $\tilde{y} = \tilde{f}(\bm x)$, and formulate the reward as:

\begin{equation}
r(\tilde{f}(\cdot)|\bm x,y) = \frac{1}{1+\mathrm{NMSE}(y, \tilde{f}(\bm x))} + \lambda \exp({-\frac{l(\tilde f(\cdot))}{L}}),
\label{eq:reward}
\end{equation} 

where $l$ represents equation complexity as the sequence length in prefix notation \citep{Kamienny-E2E-symbolic-NIPS-2022,SRBench-Cava-NeurIPS-2021,Biggio-NeSymReS-ICML-2021}; $L$ denotes the model's maximum sequence length; and $\lambda$ is a hyperparameter balancing fitting and complexity reward. Higher $\lambda$ values favor less complex equations, encouraging best-fitting and penalizing non-parsimonious solutions. NMSE is calculated as $(\frac{1}{n} \big\| y - \tilde f (\bm x) \big\|_2^2)/(\frac{1}{n} \big\| y \big\|_2^2 + \epsilon)$, where $\epsilon$ is a small constant to prevent numerical instability.

\begin{wrapfigure}[14]{r}{0.48\linewidth}
\vspace{-1.0em}
\centering
\includegraphics[width=\linewidth]{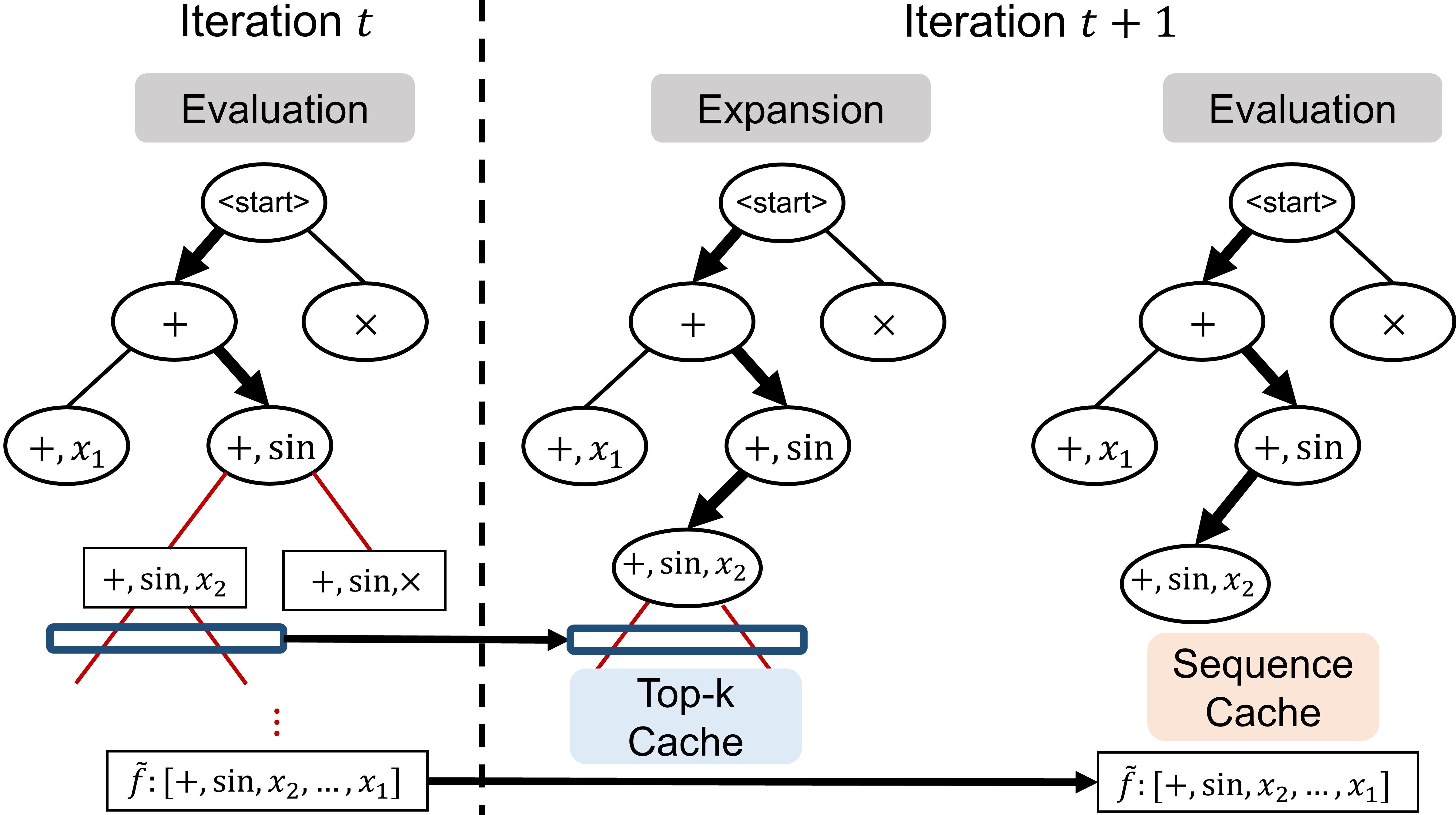}
\caption{An illustration of caching mechanisms in \modelname.
}
\label{fig:caching}
\end{wrapfigure} 
\subsection{Efficient Implementation with Caching}
During MCTS evaluation, the transformer model generates complete sequences from a given state, constructing implicit tree structures for beam search and computing $top$-$k$ next tokens for visited states. These computations are required in future MCTS iterations, so we employ two caching mechanisms, \textit{$top$-$k$ caching} and \textit{sequence caching}, to reduce redundancy and improve efficiency. 
\textit{$Top$-$k$ caching} stores computed $top$-$k$ values for given states. For example, in Fig.~\ref{fig:caching}, when evaluating state $s=[+,\mathrm{sin}]$ in MCTS iteration $t$, $top$-$k$ tokens are computed for $s$ and subsequent visited states, such as $[+,\mathrm{sin},x_2]$. State-$top$-$k$ value pairs are cached for future use, avoiding redundant token retrieval. \textit{Sequence caching} caches complete equations generated with the provided beam size. 
If a state matches a stored equation partially, the cached equation can be used directly in future iterations, bypassing iterative sequence generation. Both caching strategies enhance efficiency without compromising performance. More details are provided in Appendix \ref{sec:app_methods}.



\vspace{-0.5em}
\section{Experiments}
\label{sec:exp}
\vspace{-0.5em}
In this section, we present our experimental results that evaluate the effectiveness and efficiency of \modelname. While the proposed decoding strategy is generally model-agnostic, 
here we showcase the results of using \modelname for the end-to-end (E2E) pre-trained SR transformer backbone \cite{Kamienny-E2E-symbolic-NIPS-2022}, 
as E2E is the state-of-the-art open-source pre-trained SR model with publicly accessible model weights. Additional results of using \modelname with the NeSymReS pre-trained SR backbone \cite{Biggio-NeSymReS-ICML-2021} can be found in Appendix \ref{sec:app-tpsr_nesymres}.
 We evaluate our framework by answering the following research questions (\textbf{RQs}):
\begin{itemize}
\item[\textbf{RQ1.}]  Does \modelname perform better than other decoding strategies (beam search/sampling) and competing baseline methods over standard SR benchmark datasets?
\item[\textbf{RQ2.}] Does \modelname provide better extrapolation and robustness to noise?
\item[\textbf{RQ3.}] Are \modelname's caching mechanisms effective in reducing computation time?
\item[\textbf{RQ4.}] What is the role of individual MCTS components in \modelname's overall performance gain?
\end{itemize}

\vspace{-0.5em}
\subsection{Datasets}
\label{sec:exp_data}
\vspace{-0.5em}
We evaluate \modelname and various baseline methods on standard SR benchmark datasets from Penn Machine Learning Benchmark (PMLB) \cite{Olson2017PMLB} studied in SRBench \cite{SRBench-Cava-NeurIPS-2021}, as well as \textit{In-domain Synthetic Data} generated based on \cite{Kamienny-E2E-symbolic-NIPS-2022}. The benchmark datasets include 119 equations from \textit{Feynman Lectures on Physics database} series\footnote{\url{https://space.mit.edu/home/tegmark/aifeynman.html}} \cite{AI-Feynman2_NeuRIPS2020}, $14$ symbolic regression problems from the \textit{ODE-Strogatz database}\footnote{\url{https://github.com/lacava/ode-strogatz}} \cite{LACAVA_strogatz2016}, and 57 \textit{Black-box}\footnote{\url{https://github.com/EpistasisLab/pmlb/tree/master/datasets}} regression problems without known underlying equations. We limit the datasets to those with continuous features and input dimension $d\leq10$, as the transformer SR model \cite{Kamienny-E2E-symbolic-NIPS-2022} is pre-trained with $d_{max}=10$. The \textit{In-domain Synthetic Data} consists of $400$ validation functions with different levels of difficulty and number of input points. This data is referred to as "in-domain" because the validation functions and input points are generated using the same approach as the data on which the backbone transformer model \cite{Kamienny-E2E-symbolic-NIPS-2022} is pre-trained. More details on each of these datasets are provided in Appendix~\ref{sec:app_datasets}.

\vspace{-0.5em}
\subsection{Evaluation Metrics}
\vspace{-0.5em}
We evaluate our model using the following three metrics: $R^2$ score \cite{SRBench-Cava-NeurIPS-2021}, accuracy to tolerance $\omega$ \cite{Biggio-NeSymReS-ICML-2021,d-ascoli_icml22}, and equation complexity \cite{Kamienny-E2E-symbolic-NIPS-2022,SRBench-Cava-NeurIPS-2021}.

\vspace{-1.0em}
$$ R^2 = 1 - \frac{\sum_i^{N_{test}} (y_i - \tilde{y}_i)^2}{\sum_i^{N_{test}} (y_i - \bar{y})^2 }, \quad  \mathrm{\textit{Acc}}_{\omega} = \mathbbm{1}( \max_{1\leq i\leq N_{test}} \left|\frac{\tilde{y}_i - y_i}{y_i}\right| \leq \omega ), \quad \mathrm{\textit{Complexity}} = \left| \mathcal{T}(\tilde{f}(\cdot)) \right|,$$ 

where $R^2$ measures fitting performance with $\bar{y}$ as the mean of $y$ in test set, $\mathrm{\textit{Acc}}_{\omega}$ evaluates equation precision based on tolerance threshold $\omega$, and equation complexity is determined by the number of nodes in the expression tree $\mathcal{T}$ of the generated equation $\tilde{f}(\cdot)$. Following \cite{Kamienny-E2E-symbolic-NIPS-2022,SRBench-Cava-NeurIPS-2021}, we set $R^2=0$ for rare pathological examples and discard the worst $5\%$ predictions for $\mathrm{\textit{Acc}}_{\omega}$ to reduce outlier sensitivity.


\begin{table}[t]
\centering
\caption{\label{table-PMLB} Performance of \modelname compared with beam search and sampling decoding strategies on the SRBench \cite{SRBench-Cava-NeurIPS-2021} and In-domain Synthetic \cite{Kamienny-E2E-symbolic-NIPS-2022} datasets.
}
\resizebox{0.9\columnwidth}{!}{
\begin{tabular}{llcccccc}
\toprule
\multirow{2}{*}{Data Group    }&\multirow{2}{*}{Model}& \multicolumn{2}{c}{Feynman} & \multicolumn{2}{c}{Strogatz} & \multicolumn{2}{c}{Black-box} \\
\cmidrule(lr){3-4} \cmidrule(lr){5-6} \cmidrule(lr){7-8}
&&$\uparrow$  $R^2>0.99$   &  $\downarrow$ \textit{Complexity}  &  $\uparrow$ $R^2>0.99$   &  $\downarrow$ \textit{Complexity}  &$\uparrow$ $R^2$   &$\downarrow$  \textit{Complexity }  \\
\midrule
\multirow{6}{*}{SRBench}&E2E+Beam     & 0.815 & 54.19 & 0.357 & 53.21 & 0.847 & 83.61 \\
&E2E+Sampling & 0.848 & 50.73 & 0.357 & 50.14 & 0.864 & 82.78 \\
\cmidrule(lr){2-8}
&TPSR ($\lambda$=0) &  \textbf{0.952} &  84.42 & \textbf{0.928} & 82.78 & 0.938 & 129.85 \\
&TPSR ($\lambda$=0.1) &  0.949 &  57.22 & 0.785 & 56.14 & \textbf{0.945} &  95.71 \\
&TPSR ($\lambda$=0.5) &  0.924 &  50.01 & 0.714 & 47.02 & 0.931 &  82.58 \\
&TPSR ($\lambda$=1) &  0.916 &  \textbf{47.24} & 0.571 & \textbf{43.42} & 0.924 & \textbf{79.43} \\
\arrayrulecolor{black}\bottomrule
\\
\toprule
Data Group & Model              & $\uparrow$ $R^2>0.99$ & $\uparrow$ $R^2$ & $\uparrow$ $\mathrm{\textit{Acc}}_{0.1}$ & $\uparrow$ $\mathrm{\textit{Acc}}_{0.01}$ & $\uparrow$ $\mathrm{\textit{Acc}}_{0.001}$ & $\downarrow$ \textit{Complexity} \\
\midrule
\multirow{6}{*}{In-domain}&E2E+Beam & 0.657 &0.782 & 0.461 & 0.298  &  0.2 & 38.37 \\
&E2E+Sampling & 0.640 & 0.794 &  0.472 & 0.332 & 0.208 & 39.82\\
\cmidrule(lr){2-8}
&TPSR ($\lambda$=0) & 0.702 &0.828 &\textbf{0.550}& \textbf{0.416} &\textbf{0.333}&67.11\\
&TPSR ($\lambda$=0.1)  &\textbf{0.708}&\textbf{0.833}&0.514&0.326&0.213&40.31\\
&TPSR ($\lambda$=0.5)  &0.697&0.830&0.459&0.274&0.184&36.55\\
&TPSR ($\lambda$=1) &0.691&0.827&0.439&0.271&0.176&\textbf{35.67}\\
\arrayrulecolor{black}\bottomrule
\arrayrulecolor{black}\bottomrule
\end{tabular}
}
\end{table}





\vspace{-0.5em}
\subsection{(RQ1) Effectiveness of \modelname }
\vspace{-0.5em}
Table \ref{table-PMLB} presents the performance comparison results of \modelname with the baseline decoding strategies on the SRBench benchmark and the In-domain synthetic dataset. For the E2E baseline, we use the settings reported in \cite{Kamienny-E2E-symbolic-NIPS-2022}, including beam/sample size of $C=10$ candidates, and the refinement of all the candidates $K=10$. For our model, we use the width of tree search as $k_{max}=3$, number of rollouts $r=3$, and simulation beam size $b=1$ as the default setting. For PMLB datasets that contain more than 200 points, we follow \cite{Kamienny-E2E-symbolic-NIPS-2022} and use $B$ bags of data, each containing $N=200$ points, due to the limitation that the baseline method is pre-trained with $N\leq 200$ data points. In the baseline method \cite{Kamienny-E2E-symbolic-NIPS-2022}, a total of $BC$ candidates are generated ($C$ candidates for $B$ bags), which are then sorted and refined to generate the best equation. However, for \modelname, since we need to train an MCTS for each bag, we use an iterative decoding approach, starting with the first bag and continuing with subsequent bags until a criterion ($R^2 >0.99$) is met or we use a maximum of $B=10$ bags. To ensure a fair comparison, we use $B=10$ for the E2E baseline method as well. In this table, we demonstrate the results of our proposed framework, \modelname, with varying values of the $\lambda$ parameter that controls the trade-off between fitting performance and complexity in the hybrid reward function defined in Eq. \eqref{eq:reward}. For a detailed comparison of the experimental settings across different approaches, refer to Table \ref{table-tpsre2e-sett} in Appendix \ref{sec:app_implementation}.



\begin{figure}[t]
\centering
\includegraphics[width=0.8\linewidth]{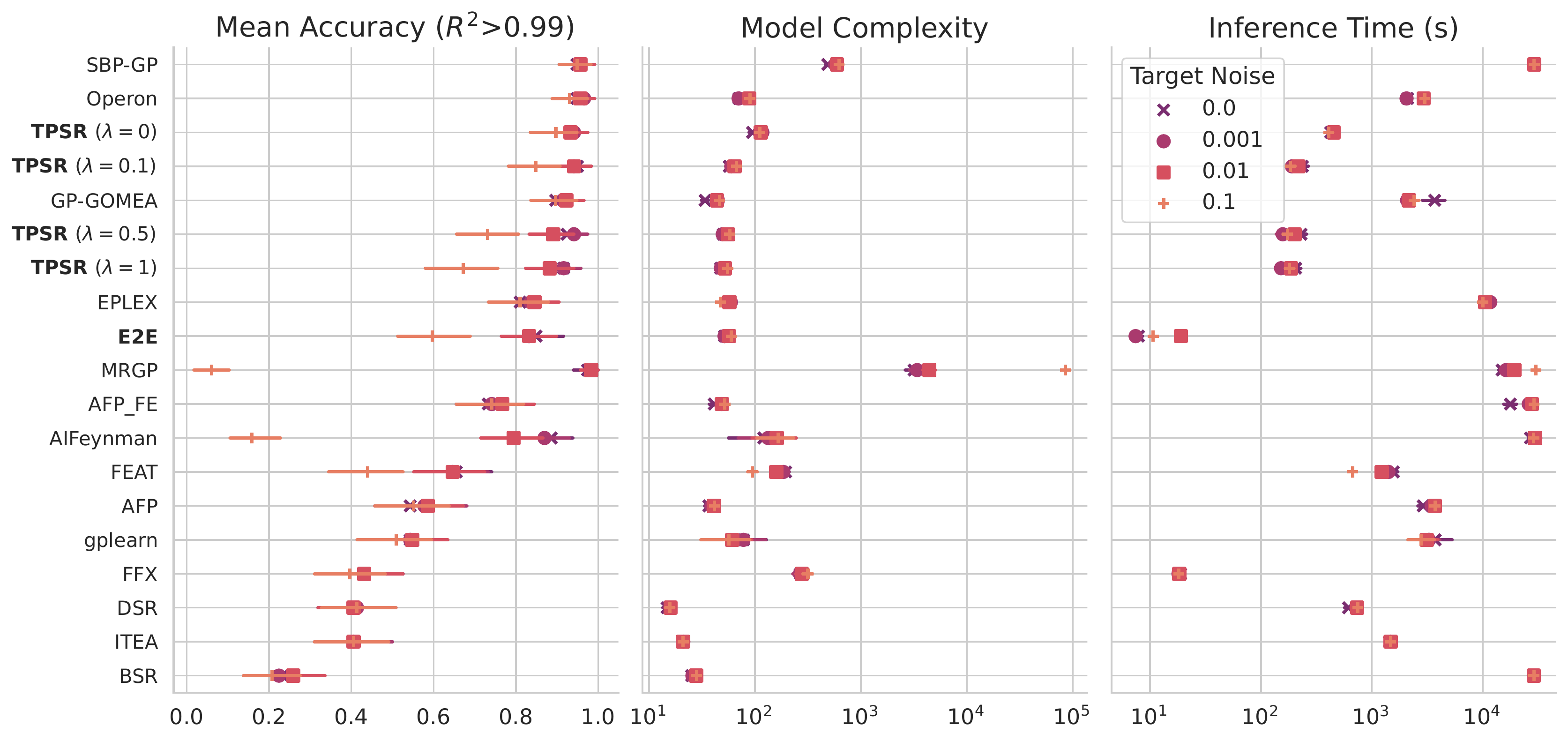} 
\includegraphics[width=0.8\linewidth]{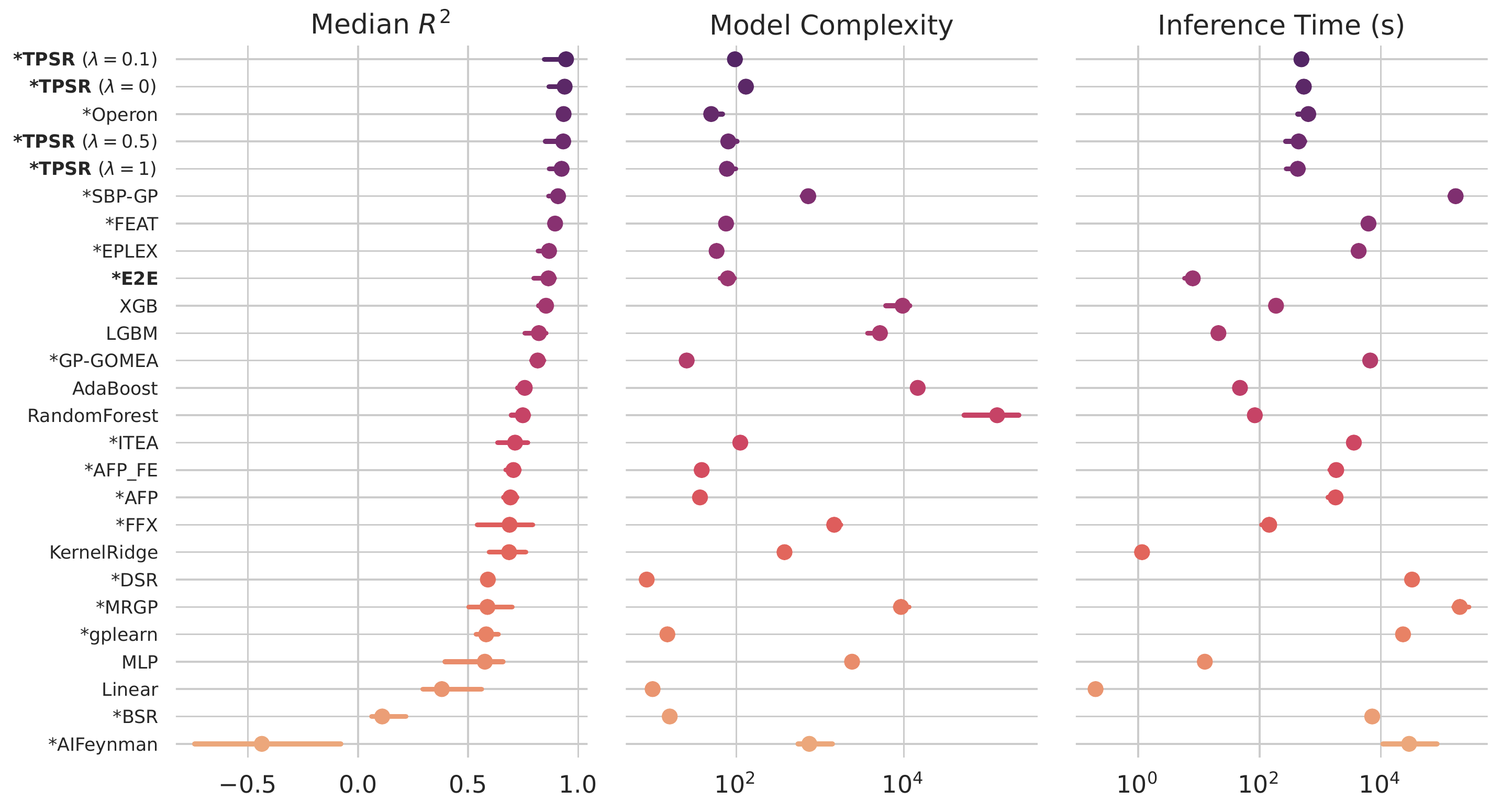}
\caption{Performance comparison of \modelname and SRBench algorithms in terms of Accuracy-Complexity-Time on \textit{Feynman} (top) and \textit{Black-box} (bottom) datasets. For \textit{Feynman} dataset, algorithms are sorted based on mean accuracy defined as the ratio of solutions with $R^2>0.99$ on test set under various noise levels, and for \textit{Black-box} dataset, the algorithms are sorted based on the median $R^2$ score on test set. TPSR demonstrates a strong balance of performance with relatively low model complexity and lower inference time compared to GP-based algorithms. The error bars represent the 95\% confidence interval and "$*$" refers to SR methods for \textit{Black-box} dataset.
}
\vspace{-0.5em}
\label{fig:feynman_parigrid}
\end{figure}

As shown in Table \ref{table-PMLB}, when $\lambda=0$, the framework generates complex equations that overoptimize for fitting performance. However, as we increase $\lambda$, the framework generates less complex equations with a slight reduction in fitting performance. Notably, even for large values of $\lambda$, such as $\lambda=1$, the fitting performance of \modelname significantly outperforms that of the baseline methods. Based on the results, we recommend a default setting of $\lambda=0.1$ as it offers a balanced trade-off between complexity and accuracy, while also mitigating potential overfitting (as detailed in Appendix \ref{sec:app-res-lambda}).
These findings demonstrate the superiority of \modelname over the baseline methods in terms of fitting performance across all datasets, while generating equations with comparable or reduced complexity than those generated by the baseline methods. Table \ref{table-PMLB} shows that \modelname exhibits a more significant gap in fitting performance when compared to E2E baselines on SRBench datasets, while this gap is smaller for In-domain datasets (even performing slightly worse on $\mathrm{\textit{Acc}}_\omega$ for larger $\lambda=0.5,1$). This is due to the In-domain dataset being generated using the same approach as the E2E pre-training data, resulting in the E2E model's superior performance on this synthetic dataset. Furthermore, qualitative comparisons of \modelname with baseline symbolic and black-box regression models \cite{XGBoost} demonstrate the superior performance of \modelname in learning the underlying equation and out-of-domain extrapolation (see Appendix \ref{sec:app-res-qual}).

Fig.~\ref{fig:feynman_parigrid} presents a detailed comparison of our proposed \modelname with the baseline E2E transformer model and all the SRBench baselines on the PMLB \textit{Feynman} and \textit{Black-box} datasets. This figure illustrate the relative position of each algorithm with respect to (1) fitting performance, (2) model complexity, and (3) inference time. The results indicate that transformer-based planning in the \modelname significantly enhances the performance of E2E and outperforms even the state-of-the-art GP baselines, achieving the highest fitting performance on the black-box datasets. This is achieved while the complexity of the generated equations in \modelname is not greater than that of E2E, and shows a great fitting-complexity balance compared to other SR algorithms. The pareto plots provided in Fig. \ref{fig:pareto-plots} also demonstrate the effectiveness of \modelname in balancing fitting and complexity compared to all other SRBench baselines. Our \modelname effectively pushes this balanced performance to the first pareto front for both the \textit{Feynman} and \textit{Black-box} datasets.
Moreover, it is important to note that, while the inference time of \modelname is longer than the baseline E2E transformer model, it still has significantly lower inference time than RL or GP-based SRBench baselines. Further results on the SRBench and In-domain datasets are provided in Appendix \ref{sec:app_results}.

\vspace{-0.5em}

\begin{figure}[t]
\centering
\includegraphics[width=0.8\linewidth]{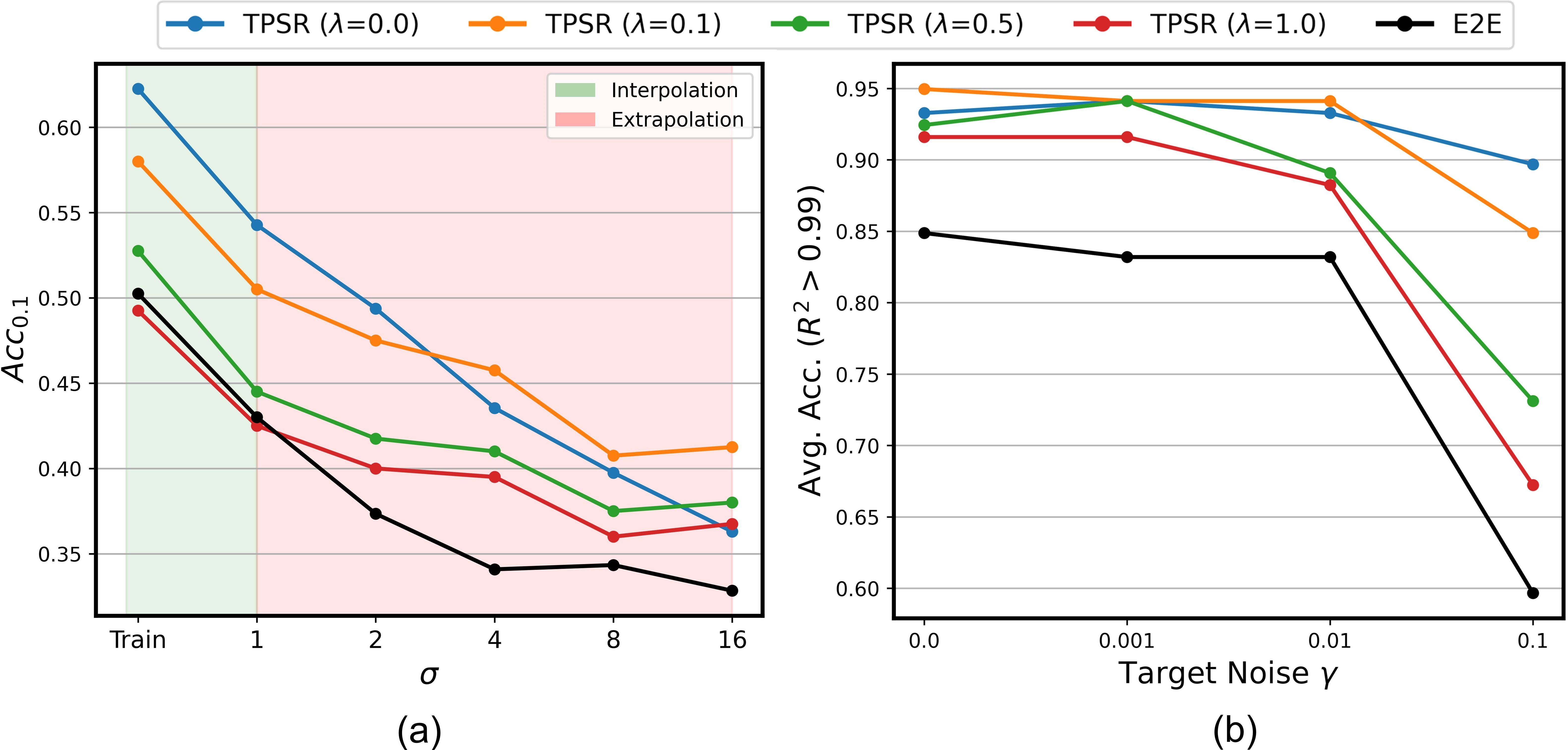}
\caption{ \modelname with $\lambda \in \{0, 0.1, 0.5, 1 \}$ compared to E2E for \textbf{(a)~Extrapolation performance} where in-domain accuracy is shown for different input variances ($\sigma$), and \textbf{(b)~Robustness to noise}, where mean accuracy ($R^2>0.99$) is shown for various target noise levels ($\gamma$).}
\label{fig:extrapolate}
\vspace{-1.0em}
\end{figure}

\subsection{(RQ2) Extrapolation and Robustness}
\vspace{-0.5em}
The ability to extrapolate well is inherently linked to the quality of the equation obtained through symbolic regression. To investigate the extrapolation performance of \modelname to out-of-training regions, we normalize the input test data points to different scales ($\sigma$) instead of unit variance (used for training points) as per \cite{Kamienny-E2E-symbolic-NIPS-2022}. Fig.~\ref{fig:extrapolate}(a) depicts the average performance of \modelname compared to E2E with sampling decoding on the training data as well as testing data in scales of $\sigma \in \{ 1, 2, 4, 8, 16 \}$ for the \textit{In-domain Synthetic} dataset. Also, we investigate the effect of different complexity controlling levels ($\lambda \in \{0, 0.1, 0.5, 1.0\}$) on the extrapolation performance. 
It can be observed that, while $\lambda=0$ (i.e., no complexity regularization) achieves the best fitting accuracy on the training data, it has a sub-par performance for $\sigma > 8$. This can be due to the overfitting issue when the symbolic model is much more complex than the real complexity of the equation, similar to the common overfitting issue in ML models. 
The results highlight the importance of controlling complexity in the extrapolation of identified equations. For values of $\lambda > 0$, the overfitting issue is mitigated as the generated equations become less complex. However, very high values of $\lambda$ (e.g., $\lambda=1$) can result in poor fitting performance. The flexibility of \modelname for allowing different values of $\lambda$ to balance fitting and complexity for a given task is crucial for optimal performance. 
Fig.~\ref{fig:extrapolate}(b) also presents the robustness of \modelname with different $\lambda$ levels compared to the E2E transformer baseline on the \textit{Feynman} dataset. The results indicate that MCTS-guided decoding can offer robust performance with a smaller drop compared to the baseline in the presence of noise.

\vspace{-0.5em}
\subsection{Ablation Study}
\vspace{-0.5em}
In this section, we investigate the effect of different MCTS parameters and caching mechanisms on the performance of \modelname by conducting ablative experiments on the \textit{Feynman} datasets.

\begin{figure}[t]
\centering
\includegraphics[width=0.75\linewidth]{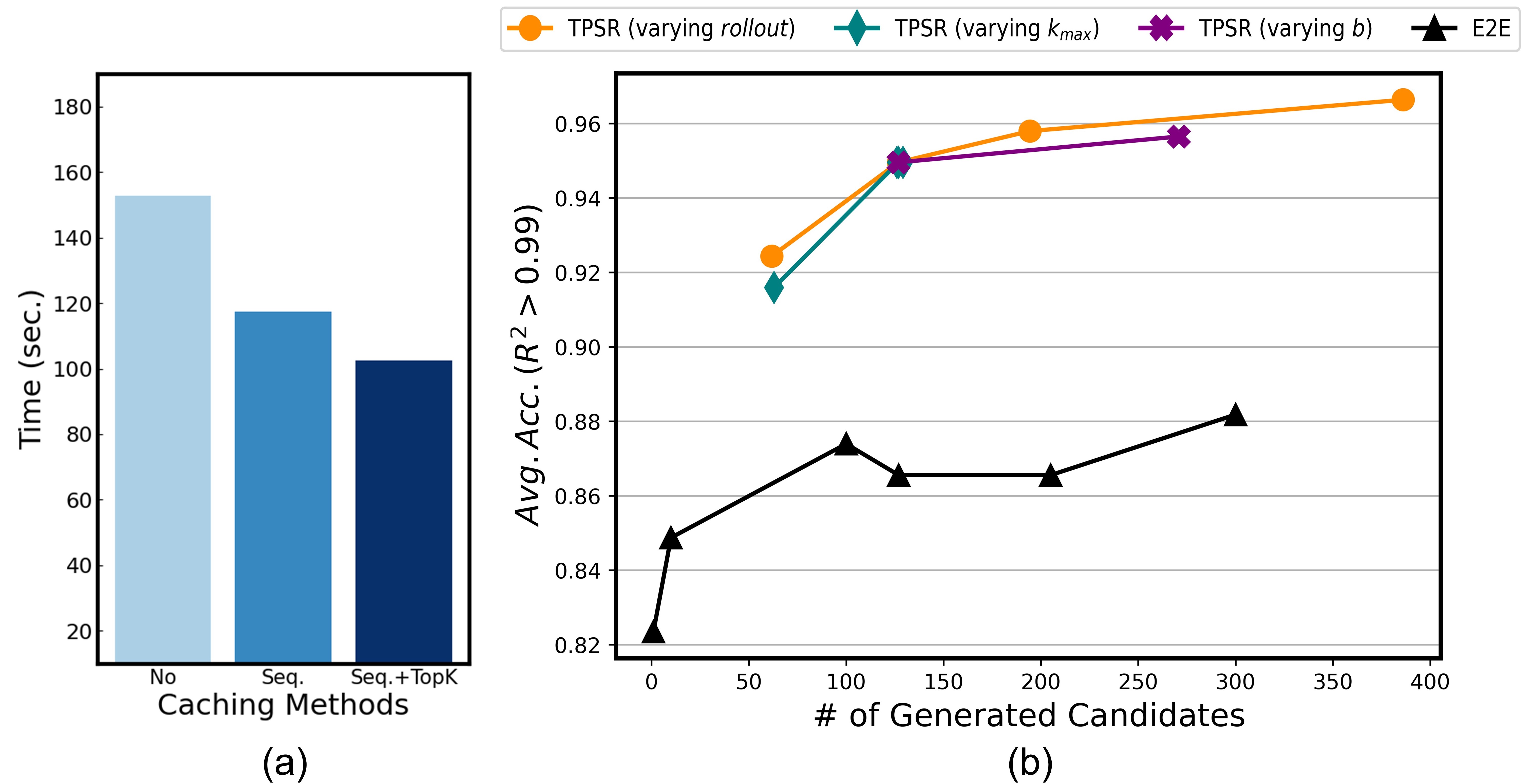} 
\caption{Ablation study on the modules and parameters of \modelname. \textbf{(a)~Effect of caching mechanisms:} \textit{Sequence caching} and \textit{$top$-$k$ caching} improve the inference time of \modelname ($\lambda=0.1$). \textbf{(b)~Efficiency and parameters of \modelname}: Average accuracy of \modelname (varying model parameters), and baseline E2E (varying sampling size) vs. number of generated candidates.
}
\label{fig:ablation}
\end{figure}
\noindent \textbf{(RQ3) Caching Mechanisms. } In Fig.~\ref{fig:ablation}(a), we illustrate the effectiveness of the \textit{sequence} and $top$-$k$ caching mechanisms in reducing the total inference time of \modelname ($\lambda=0.1$). Our experiments show that sequence caching has more effect in dropping the inference time as it replaces the time-consuming sequence generation process. Overall, these two mechanisms can reduce the total inference time by around $28\%$. 



\noindent \textbf{(RQ4) Search Parameters. }Fig.~\ref{fig:ablation}(b) shows the fitting performance against the number of generated equations during the decoding process for both \modelname ($\lambda=0.1$) and the baseline E2E with sampling decoding. In this figure, the `number of generated equation candidates' represents the total number of complete equation sequences generated by each method. Specifically, this refers to the sample size in the E2E with sampling decoding, and the function calls of the beam search sub-routine multiplied by beam size $b$ in \modelname.
The results show that under the same number of generated equation candidates, \modelname significantly outperforms the E2E baseline. This is primarily attributed to the fact that the E2E baseline is deprived of any feedback on the fitting performance of the generated equations.
We report the results for variants of \modelname with different MCTS parameters. We assess the performance with varying number of rollouts, $r=\{1,3,6,9\}$, number of beams in simulations, $b=\{1,3\}$, and the maximum number of possible expansions at each state, $k_{max}=\{2,3,4\}$. The default setting of \modelname parameters are $b=1$, $k_{max}=3$, and $r=3$. The results indicate that increasing $r$, $k_{max}$, and $b$ all contribute to the better performance of \modelname, with the most significant improvement observed when increasing $r$. This is because more rollouts provide the model with more opportunities to learn from trials and learn better values.

\vspace{-0.5em}
\section{Conclusion}
\label{sec:conc}
\vspace{-0.5em}
In this work, we propose \modelname, a model-agnostic decoding strategy for symbolic regression that leverages the power of pre-trained SR transformer models combined with MCTS lookahead planning, and outperforms the existing methods in generating equations with superior fitting-complexity trade-off. We demonstrate the flexibility of \modelname in controlling equation complexity without fine-tuning the pre-trained model. Results also show that better expressions obtained with lookahead planning can further improve model performance in terms of noise robustness and extrapolation to unseen data. 
Future research could focus on enhancing the adaptability of feedback-based expression generation mechanisms, potentially by modulating the flexibility of MCTS or SR model weights, and the integration of MCTS with the training or fine-tuning of transformer SR models.
Furthermore, employing parallelization and distributed computing could potentially improve MCTS search efficiency.

\newpage

\small

\bibliographystyle{unsrt}
\bibliography{neurips_2023.bbl}

\normalsize

\newpage 

\appendix 

\section*{Appendix}
\vspace{-0.5em}
\section{Dataset Details}
\label{sec:app_datasets}
\vspace{-0.5em}
We evaluate \modelname and several baseline methods on the following four standard benchmark datasets: \textit{Feynman}, \textit{Black-box}, and \textit{Strogatz} from SRBench \cite{SRBench-Cava-NeurIPS-2021}, and \textit{In-domain Synthetic Data} generated based on \cite{Kamienny-E2E-symbolic-NIPS-2022}. More details on each of these datasets are given below.

\begin{myitemize2}
\item \textbf{\textit{Feynman\footnote{\url{https://space.mit.edu/home/tegmark/aifeynman.html}}}: }
\label{sec:exp_data_aif}
This dataset contains $119$ equations sourced from \textit{Feynman Lectures on Physics database} series \cite{AI-Feynman2_NeuRIPS2020}. The regression input points $(x,y)$ from these equations are provided in Penn Machine Learning Benchmark (PMLB) dataset \cite{SRBench-Cava-NeurIPS-2021, Olson2017PMLB} and have been studied in SRBench \cite{SRBench-Cava-NeurIPS-2021} for the symbolic regression task. The input dimension is limited to $d\leq 10$ and the true underlying function of points is known. We split the dataset into $B$ bags of $200$ input points (when $N$ is larger than $200$) since the transformer SR model is pretrained on $N \leq 200$ input points as per \cite{Kamienny-E2E-symbolic-NIPS-2022}.
\item \textbf{\textit{Strogatz\footnote{\url{https://github.com/lacava/ode-strogatz}}}:} This dataset comprises $14$ symbolic regression problems sourced from the \textit{ODE-Strogatz database} \cite{LACAVA_strogatz2016} for nonlinear dynamical systems. The input points for these problems are included in PMLB \cite{Olson2017PMLB} and have been examined in SRBench \cite{SRBench-Cava-NeurIPS-2021} for symbolic regression. The input dimension for these problems is restricted to $d=2$ and the true underlying functions are provided.
\item \textbf{\textit{Black-box\footnote{\url{https://github.com/EpistasisLab/pmlb/tree/master/datasets}}}:} The black-box regression datasets from PMLB \cite{Olson2017PMLB} are used for the symbolic regression task and studied in SRBench \cite{SRBench-Cava-NeurIPS-2021} among various baselines. The aim of SR study on these black-box datasets is to find an interpretable model expression that fits the data effectively. We limit the datasets to those with continuous features and input dimension $d\leq10$, as the transformer SR model \cite{Kamienny-E2E-symbolic-NIPS-2022} is pretrained with $d_{max}=10$. In total, there are $57$ black-box datasets that consist of real-world and synthetic datasets with varying levels of noise. 
\item \textbf{\textit{In-domain Synthetic Data}:} Following \cite{Kamienny-E2E-symbolic-NIPS-2022}, we construct a fixed validation set consisting of $400$ equation examples in which the validation functions were uniformly distributed across three different difficulty factors: input dimension ($d$), number of unary  operators ($u$), and binary operators ($b$). Specifically, we set $d \sim \mathcal{U}(1,d_{max})$, $b \in \mathcal{U}(d-1,d+b_{max})$, and $u \sim \mathcal{U}(0,u_{max})$, where $d_{max}=10$, $u_{max}=5$, and $b_{max}=5+d$. The equation sequence is generated for each function by providing $N=[50,100,150,200]$ input points $(x,y)$, and the prediction accuracy is assessed on $N_{test}=200$ points that are randomly extracted from a multimodal distribution, as described in \cite{Kamienny-E2E-symbolic-NIPS-2022}. This data is referred to as ``in-domain'' because the validation data is generated using the same approach as the data on which the model \cite{Kamienny-E2E-symbolic-NIPS-2022} is pretrained. 
\end{myitemize2}

\vspace{-0.5em}
\section{Implementation Details}
\label{sec:app_implementation}
\vspace{-0.5em}
Our model implementation leverages the state-of-the-art open-source End-to-End (E2E) SR model \cite{Kamienny-E2E-symbolic-NIPS-2022} as the pre-trained transformer backbone. This selection is due to the public availability of E2E's model architecture, weights, and logits in the Facebook \texttt{symbolicregression} library \footnote{\url{https://dl.fbaipublicfiles.com/symbolicregression/}} and repository \footnote{\url{https://github.com/facebookresearch/symbolicregression}}. The algorithm of our model is provided in Appendix \ref{sec:app_methods} and the implementation code for our experiments with configuration details for reproducibility is available \footnote{\url{https://github.com/deep-symbolic-mathematics/TPSR}}.
In our experiments, the model's maximum sequence length is set to $L=200$, and the constant to prevent numerical stability $\epsilon$ in NMSE calculation $(\frac{1}{n} \| y - \tilde f (\bm x) \|_2^2)/(\frac{1}{n} \| y \|_2^2 + \epsilon)$ is set to $1e-9$. We set the default maximum number of node expansions ($k_{max}$) to be 3, the beam size of simulations ($b$) as 1, and the number of rollouts ($r$) as 3. The complexity-controlling parameter ($\lambda$) was also varied across four values: ${0,0.1,0.5,1}$. To ensure consistency with the protocol set out by \cite{Kamienny-E2E-symbolic-NIPS-2022}, we divided the observation points of each equation in the SRBench datasets (including \textit{Feynman}, \textit{Strogatz}, and \textit{Black-box}) into training and testing sets at a ratio of $75\%/25\%$. In the evaluation experiments involving \textit{In-domain Synthetic Data}, we adjusted the number of observation points for each equation on which TPSR was trained to $N\in[50,100,150,200]$. The generated expression was subsequently tested on the $N_{test}=200$ data points for each sampling variance ($\sigma$) of 1, 2, 4, 8, and 16. These synthetic input points with varying sampling variance are introduced in \textit{In-domain} data \cite{Kamienny-E2E-symbolic-NIPS-2022} to assess the models' extrapolation capabilities under different conditions. All experiments are implemented with PyTorch on four Quadro RTX 8000 GPUs, with 48GB of RAM. 

\begin{table}[t]
\centering
\caption{Experimental Settings of \modelname and E2E \cite{Kamienny-E2E-symbolic-NIPS-2022}}
\resizebox{0.55\textwidth}{!}{
\begin{tabular}{ccc}
\toprule
Setting/Parameter & \modelname & E2E \\
\midrule
Maximum Equation Length ($L$) & $200$ & $200$\\
Maximum No. of Observations ($N$) & $200$ & $200$\\
Maximum Input Dimension ($D$) & $10$ & $10$\\
Maximum No. of Bags ($B$) & $10$ & $10$\\
Beam/Sample size ($C$) & \textendash  & $10$ \\
No. of Refinement Candidates ($K$) & \textendash & $10$ \\
Maximum Expansion Width ($k_{max}$)  & $3$ & \textendash  \\
Maximum No. of Rollouts ($r$) & $3$ & \textendash \\
Beam Size in Simulations ($b$) & $1$ & \textendash \\
UCT Exploration Parameter ($\beta$)  & $1$ & \textendash  \\
\bottomrule
\end{tabular}
}
\label{table-tpsre2e-sett}
\vspace{-1.0em}
\end{table}

\vspace{-0.5em}
\section{Methodology Details}
\label{sec:app_methods}
\vspace{-0.5em}
\subsection{MCTS-Guided Decoding Details}
\vspace{-0.5em}
Algorithm \ref{alg:tpsr} provides the details of steps in MCTS-Guided decoding strategy for SR, following \cite{MCTS-code-ICLR-2023}. Here, the \textcolor{blue}{blue} lines correspond to the utilization of reward and selection functions defined in Eqs. \eqref{eq:reward} and \eqref{eq:uct} of Section 3. These functions play a crucial role in guiding the MCTS-based Transformer Decoding strategy for SR and ensuring effective exploration and exploitation within the search space. Meanwhile, the \textcolor{purple}{red} lines in the algorithm denote the places when the pre-trained transformer SR model is invoked to extract the $top$-$k$ next tokens and equation candidate beams. These extracted tokens and beams are employed in the expansion and evaluation steps of the MCTS algorithm, respectively. By incorporating the pre-trained transformer SR model, the MCTS-Guided decoding strategy can effectively leverage the model's inherent semantic knowledge gained through large-scale pre-training to generate high-quality equation candidates and enhance the overall performance of the SR approach. 
Notably, in this MCTS setting, a "visit" signifies that a state-action pairing, $(s,a)$, has been explored during tree search, appending the corresponding child state, $s'$, to the tree. Sequences that are generated as part of the beam search sub-routine of simulations in the evaluation stage of MCTS are not directly considered as visits to the nodes corresponding to these sequences. Instead, they serve the purpose of completing the partial equation to allow for feedback computation. As for cache hits, they are also not counted as visits. The reason is that caching in this context is used to save computation by storing previously computed values, and a cache hit simply means retrieving a stored value rather than performing a new visit.

{
\centering
\begin{minipage}{0.9\linewidth}
\begin{algorithm}[H]
\fontsize{9}{9}\selectfont
\SetAlgoLined
\SetKwInOut{Require}{Require}
\SetKwInOut{Input}{Input}
\SetKwInOut{Output}{Output}
\SetKw{KwDownTo}{downto}
\SetKw{KwAnd}{and}
\SetKw{KwContinue}{continue}
\caption{MCTS-Guided Decoding for Symbolic Regression}
\label{alg:tpsr}
\Input{ $r_{max}$: maximum number of rollouts, $k_{max}$: number of children of nodes used for \textit{top-k} next token selection, $b$: beam size, $c$: P-UCB exploration parameter}
\BlankLine
\While{$r < r_{max}$}{
    \textit{node} $\leftarrow$ \textit{root}\;
    \textcolor{brown}{1) Selection} \\
    \While{$|\textit{node.children}| > 0$}{
        node $\leftarrow$ \textcolor{blue}{SELECT}(\textit{node.children}, $c$)\;
    }
    \textcolor{brown}{2) Expansion} \\
    next tokens $\leftarrow$ \textcolor{purple}{TOP\_K}(\textit{node}, $k_{max}$)\;
    \For{\textit{action} $\in$ next tokens}{
        next state $\leftarrow$ CONCAT(\textit{node}, \textit{action})\;
        Add next state as a child of \textit{node}\;
    }
    \textcolor{brown}{3) Evaluation} \\
    Equation $\leftarrow$ \textcolor{purple}{BEAM\_SEARCH}(\textit{node}, $b$)\;
    \textit{reward} $\leftarrow$ \textcolor{blue}{GET\_REWARD}(Equation)\;
    Save (Equation, \textit{reward}) pair in a dictionary. \;
    \textcolor{brown}{4) Backpropagation} \\
    Update values on the trajectory given the \textit{reward} \;
}
Return Equation with the highest \textit{reward} \;
\end{algorithm}
\end{minipage}
\par
}

\vspace{-0.5em}
\subsection{Distinguishing TPSR from other MCTS Approaches in SR}
\vspace{-0.5em}
\begin{figure}[t]
\centering
\includegraphics[width=0.9\linewidth]{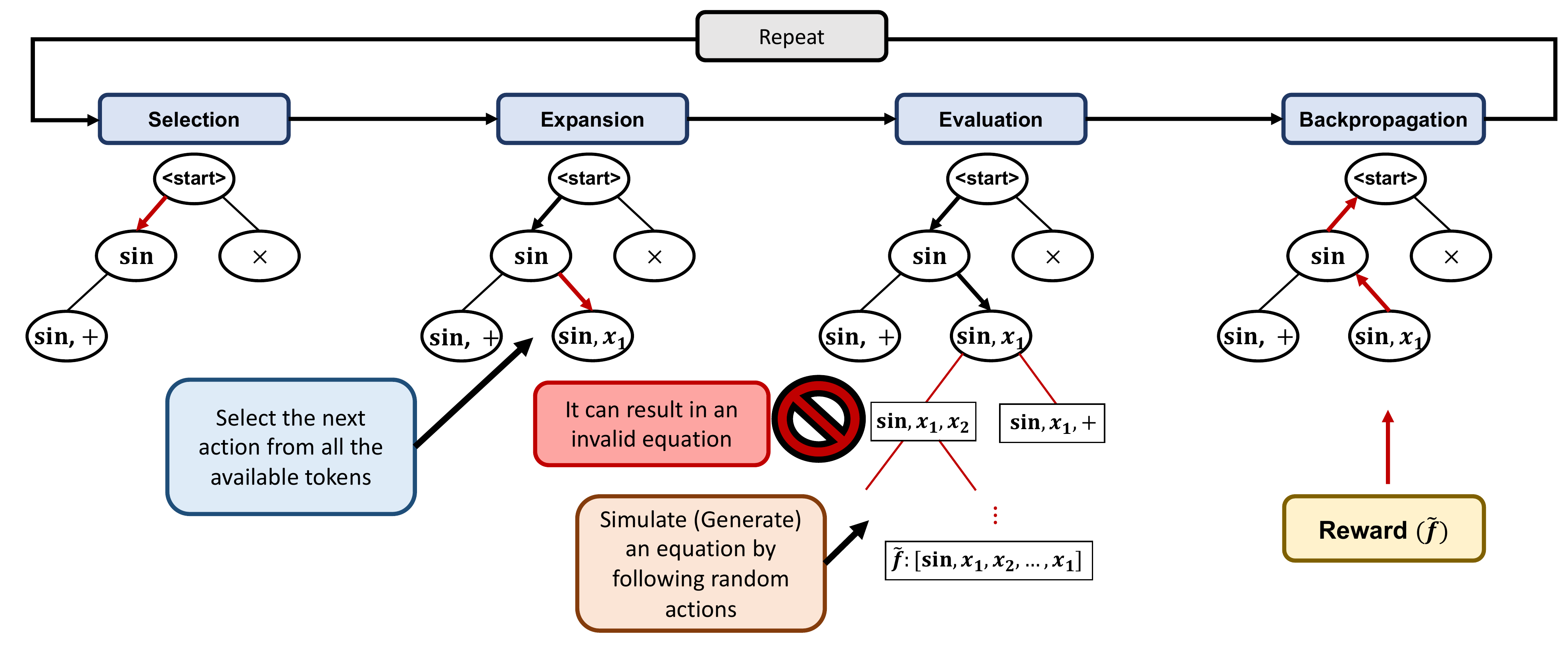}
\caption{MCTS-Guided decoding algorithm for Symbolic Regression without using the pretrained transformer SR model for expansion and evaluation steps.
}
\label{fig:mcts_wo_transformer}
\end{figure}
\begin{figure}[t]
\centering
\includegraphics[width=0.9\linewidth]{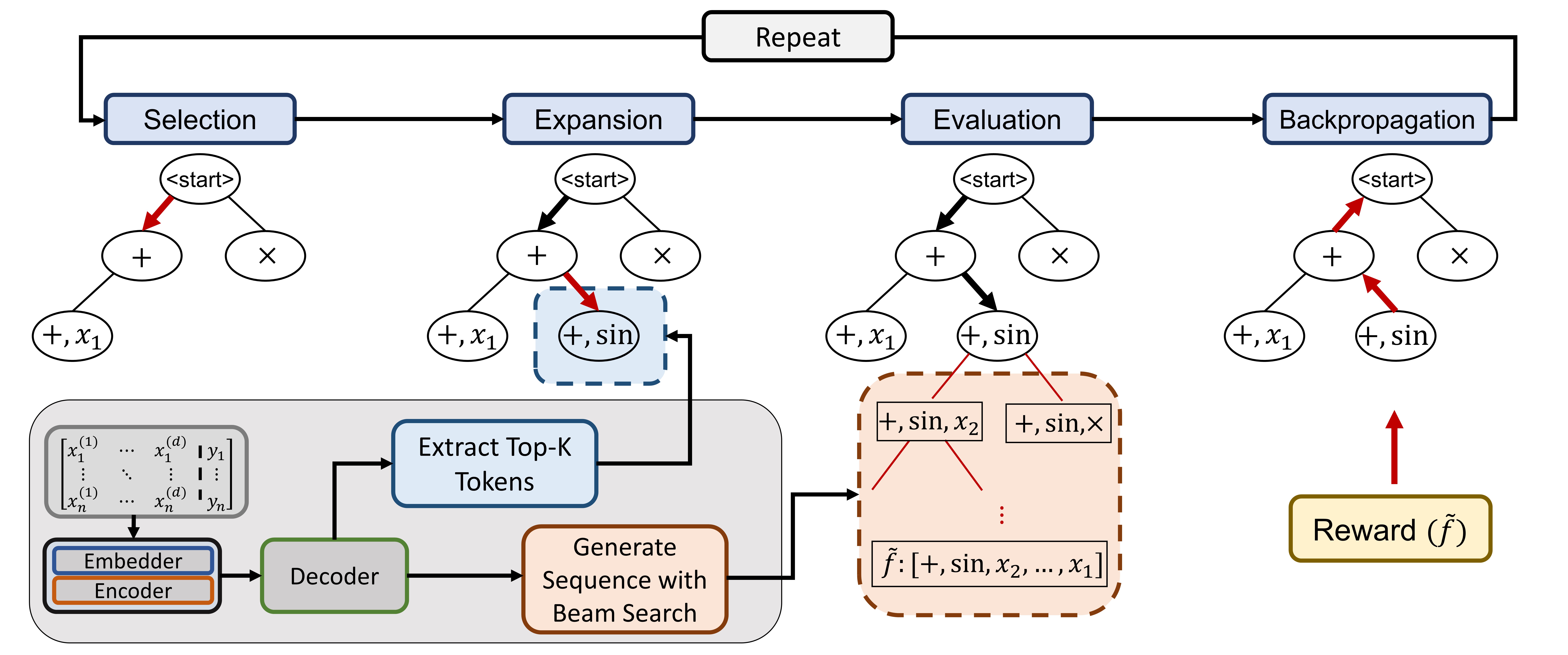}
\caption{MCTS-Guided decoding algorithm for Symbolic Regression with the pretrained Transformer model used for expansion and evaluation steps.
}
\label{fig:mcts_appendix}
\end{figure}
It is essential to highlight that the implementation of the MCTS approach in TPSR differs from the standalone MCTS algorithm for SR. In a recent work, Sun \textit{et al.} \citep{SPL-MCTS-ICLR-2023} shows that Monte Carlo Tree Search can be effective for exploring the optimal expression trees that govern nonlinear dynamical systems. This work introduces several adjustments to the conventional MCTS to enable the recovery of equations as expression trees. However, we would like to remark that using MCTS as a standalone algorithm for SR is a single-instance SR method, meaning that it requires searching from scratch for a new function or measurement data, and does not leverage pre-trained SR priors.
To highlight the role of pre-trained transformer in our \modelname framework, we compare the MCTS-guided decoding algorithm in \modelname (Fig. \ref{fig:mcts_appendix}, replicated from the main body for ease of comparison) with a standard MCTS algorithm (Fig. \ref{fig:mcts_wo_transformer}) which can be used in a similar fashion but without sharing information with the transformers. During the expansion phase, standard MCTS chooses the next accessible action from the action set (i.e., the vocabulary of tokens) and appends the state that can be reached through the chosen action. In this example, action $x_1$ is selected, and the new state appended to the tree is $[sin, x_1]$. Subsequently, during the evaluation phase, MCTS assesses the new state by implementing a \textit{random} policy from the new state and calculating the policy's value. Applying the standard MCTS algorithm to domains characterized by extensive state or action spaces, such as SR with a combinatorial optimization space that exponentially grows with the number of input variables, is highly impractical. This is because attempting all possible actions in the expansion phase is infeasible. Furthermore, the random policy employed in the evaluation phase exhibits significant variance when estimating the new state's value, and may result in invalid equations that are unsuitable for proper evaluation (e.g., accurately assessing fitting performance). To overcome these limitations, \modelname employs the pre-trained Transformer SR model. This approach leverages the semantic knowledge embedded in large-scale pre-trained priors, while conducting lookahead planning to optimize equation generation for the equation discovery non-differentiable objectives. By integrating the pre-trained Transformer SR model, \modelname can effectively navigate the vast search space, reducing complexity and enhancing fitting performance, thus offering a more viable solution for the domain of SR.

It is also crucial to emphasize how the integration of MCTS in \modelname differentiates from others, particularly from works like Kamienny \textit{et al.} \citep{pmlr-v202-kamienny23a}, which also pairs MCTS with pre-trained Transformers. 
Key differentiators include:

\textbf{General Approach.} 
Unlike \cite{pmlr-v202-kamienny23a} that exploits a pre-trained mutation policy $M$ to generate the expression by following a series of mutations from an empty expression (root), \modelname follows the seq2seq approach of E2E \cite{Kamienny-E2E-symbolic-NIPS-2022} to generate the expression token-by-token. Consequently, \modelname uses the pre-trained E2E as its backbone but \cite{pmlr-v202-kamienny23a} pre-trains the mutation policy from scratch.

\textbf{MCTS and Search Strategy.} 
In \cite{pmlr-v202-kamienny23a}, the search tree consists of full mathematical equations, with each node representing a distinct equation and edges corresponding to mutations between equations. In contrast, \modelname employs MCTS as a decoding strategy in the context of the transformer model. Each node in the search tree of \modelname represents the current state of generated tokens, potentially forming non-complete sequences, with edges corresponding to mathematical operators or variables. So, the search tree of \cite{pmlr-v202-kamienny23a} with ``n'' nodes includes ``n'' different equations, while the \modelname search tree includes partial decoded sequences, and completed equations only exist at the terminal nodes. This distinction inherently leads to major differences in selection, expansion, and back-propagation mechanisms within the MCTS algorithm.

\textbf{Parameter Update and Learning.} 
\cite{pmlr-v202-kamienny23a} utilizes MCTS to update and learn the distribution of mutations for a group of out-of-distribution datasets. The approach involves fine-tuning an actor-critic-like model to adjust the pre-trained model on a group of symbolic regression instances. On the other hand, \modelname uses the pre-trained transformer's learned distribution to guide the expansion during the search process, without updating any specific parameters for in-domain or out-of-domain datasets (without fine-tuning). Consequently, the same settings and pre-trained model are applied to both in-domain and out-of-domain evaluations in \modelname.

\textbf{Computation Time.}
\cite{pmlr-v202-kamienny23a} involves pre-training a mutation policy, a critic network, and performing fine-tuning stages for these networks, leading to significantly higher computation time (a limit of $24$hrs and $500$K equation candidate evaluations as stated in \cite{pmlr-v202-kamienny23a}). In contrast, \modelname has substantially lower computation time and the number of equation candidate evaluations, typically in the order of $10^2$ equations, taking approximately $10^2$ seconds (as shown in Fig.~\ref{fig:feynman_parigrid} and \ref{fig:ablation}). This renders \modelname more suitable for applications where fast yet accurate equation discovery is critical. 

\vspace{-0.5em}
\subsection{Caching Details}
\vspace{-0.5em}
In the evaluation phase of MCTS, a transformer model is employed to produce complete sequences from a given state. This procedure entails the creation of implicit tree structures that are used to carry out a beam search. The beam search involves determining the $top$-$k$ next tokens for the states visited during the generation process until the entire sequence is generated. These calculations will be needed in future MCTS iterations for two purposes: (1) to extract the $top$-$k$ next tokens during the \textbf{expansion} step of each state and (2) to generate the complete equation from a given state during the \textbf{evaluation} step. To avoid redundant computations and improve the efficiency of the framework, two caching mechanisms are used, namely \textit{$top$-$k$ caching} and \textit{sequence caching}.

\textit{Top-k caching} is a mechanism that stores the computed top-k values for given states. For example, in Fig.~4 of the main paper, when evaluating the state $s=[+, \mathrm{sin}]$ in iteration $t$ of MCTS, the $top$-$k$ tokens are calculated for $s$ and its subsequent visited states (e.g., $[+, \mathrm{sin}, x_2]$). These pairs of states and their corresponding $top$-$k$ values can be stored in a \textit{top-k cache}. Consequently, if a state $s$ is visited again in a future iteration (e.g., visiting $s=[+, \mathrm{sin}, x_2]$ in iteration $t+1$ of MCTS), the cached $top$-$k$ values are utilized instead of calling pretrained SR model again and retrieving the $top$-$k$ tokens from the forward pass of model.

Another mechanism employed to reduce redundant computations is \textit{sequence caching}, which caches complete equations generated in a greedy manner. When the beam size in MCTS is one, the sequence is generated greedily for the given state in the evaluation step. This means that if any partial sequence of this equation is given, the same equation will be generated by the decoder. As a result, the generated equation in iteration $t$ can be used directly in future iterations if the state matches the stored equation partially. For instance, in Fig.~4 of the main paper, consider the equation $\tilde{f}:[+, \mathrm{sin}, x_2, \cdot, x_1]$ is generated for $s=[+, \mathrm{sin}]$ with $b=1$ in iteration $t$. Now, if in a later iteration (e.g., iteration $t+1$), the state to evaluate is $s=[+, \mathrm{sin}, x_2]$, the iterative sequence generation process can be bypassed by directly using the sequence cache to predict the complete equation. It is essential to note that both of these caching strategies serve the same purpose of enhancing the framework's efficiency without compromising its performance.

\vspace{-1.0em}
\section{Further Results and Visualization}
\label{sec:app_results}
\vspace{-0.5em}
\subsection{Controlling the Fitting-Complexity Trade-off}
\label{sec:app-res-lambda}

\begin{wrapfigure}[26]{r}{0.40\linewidth}
\centering
\vspace{-1.0em}
\includegraphics[width=\linewidth,keepaspectratio]{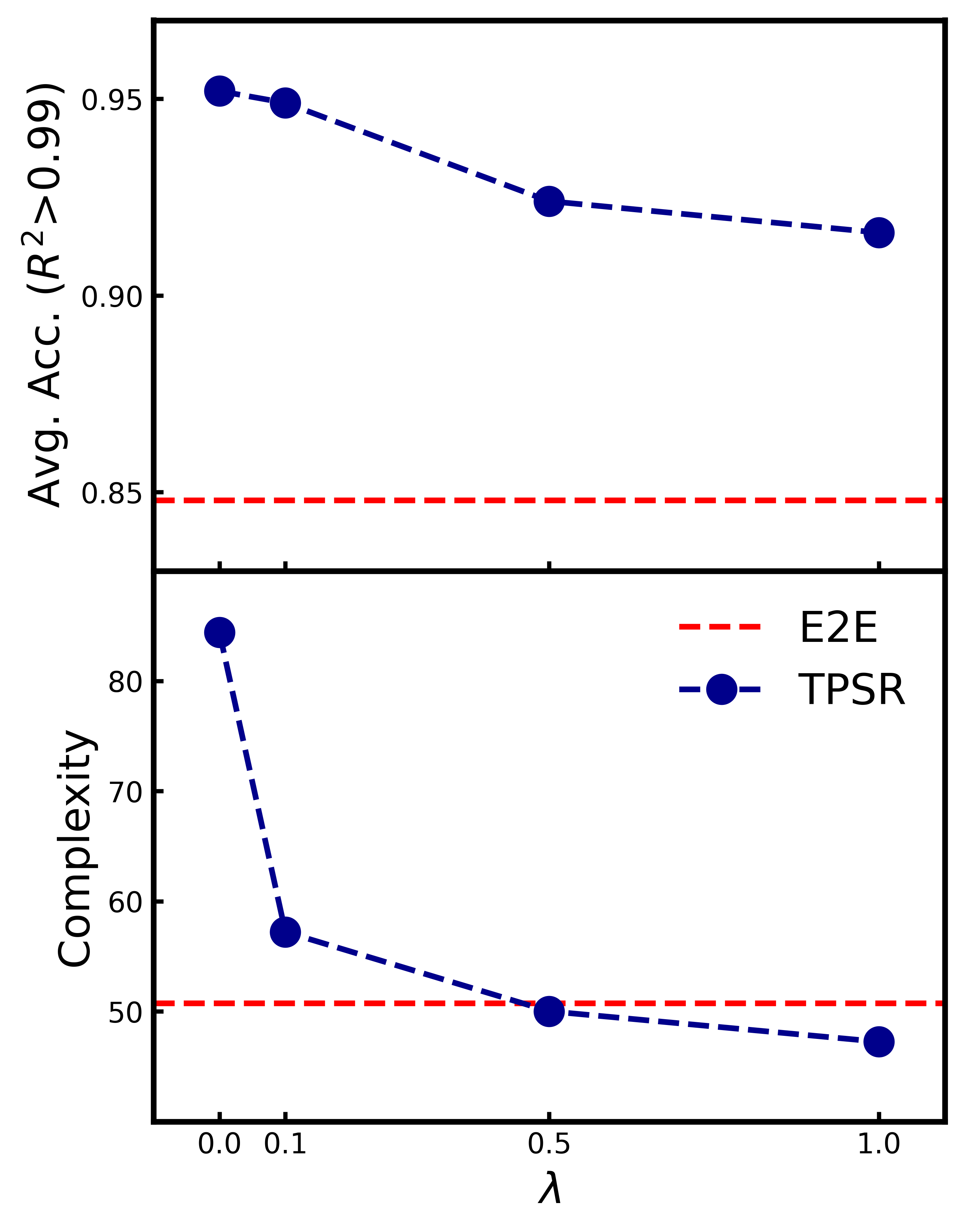}
\caption{ Effect of controllable complexity parameter ($\lambda$) on average test performance and equation complexity for the \textit{Feynman} dataset. E2E uses sampling decoding.}
\label{fig:rc}
\vspace{-1.0em}
\end{wrapfigure}

\vspace{-0.5em}
Figure~\ref{fig:rc} illustrates the relationship between fitting accuracy and complexity of predicted equations for various values of $\lambda \in \{0,1,0.5,1\}$, on the \textit{Feynman} dataset. 
This figure highlights the impact of the controllable complexity parameter $\lambda$ on balancing the trade-off between fitting performance and equation complexity.
As it can be observed, when the value of $\lambda$ is set to $0$, the TPSR framework generates exceedingly complex equations, resulting in a complexity score greater than $80$. These equations are primarily focused on optimizing fitting performance. However, as $\lambda$ is slightly increased to $0.1$, there is a minimal effect on the fitting performance, while the complexity of the generated equations drops significantly to a score of less than $60$.
As $\lambda$ continues to increase, the TPSR framework produces equations with reduced complexity, accompanied by a slight decline in fitting performance. Figure~\ref{fig:rc} demonstrates that even when $\lambda$ is set to a large value, such as $1$, the fitting performance of the equations generated by TPSR remains notably superior to the baseline E2E+Sampling method ($\bm{0.916}$ versus $\bm{0.848}$). Additionally, the complexity of the generated equations marginally improves ($\bm{47.24}$ compared to $\bm{50.73}$). This can be observed by examining the gap between the red and blue dashed lines in both the top and bottom sub-figures of Figure~\ref{fig:rc}.
These findings emphasize the advantages of the TPSR framework over the baseline methods in terms of fitting performance. At the same time, TPSR is capable of generating equations with either comparable or lower complexity than those produced by the baseline methods.

Given the significance of $\lambda$ in governing this trade-off, and to assist users in hyperparameter selection, we recommend setting $\lambda=0.1$ as a default. Based on our results, particularly Table.~\ref{table-PMLB} and Fig.~\ref{fig:rc}, we find that this setting tends to achieve a harmonious balance between accuracy and complexity, mitigating overfitting. It is important to note that this recommendation aims to offer a starting point for users. The appropriate choice of this hyperparameter may depend on the specific use case, where the balance between finding an accurate function and sacrificing complexity, versus emphasizing interpretability and equation simplicity over relative accuracy, becomes relevant. 

\begin{figure}[t]
\centering
\includegraphics[width=0.8\linewidth]{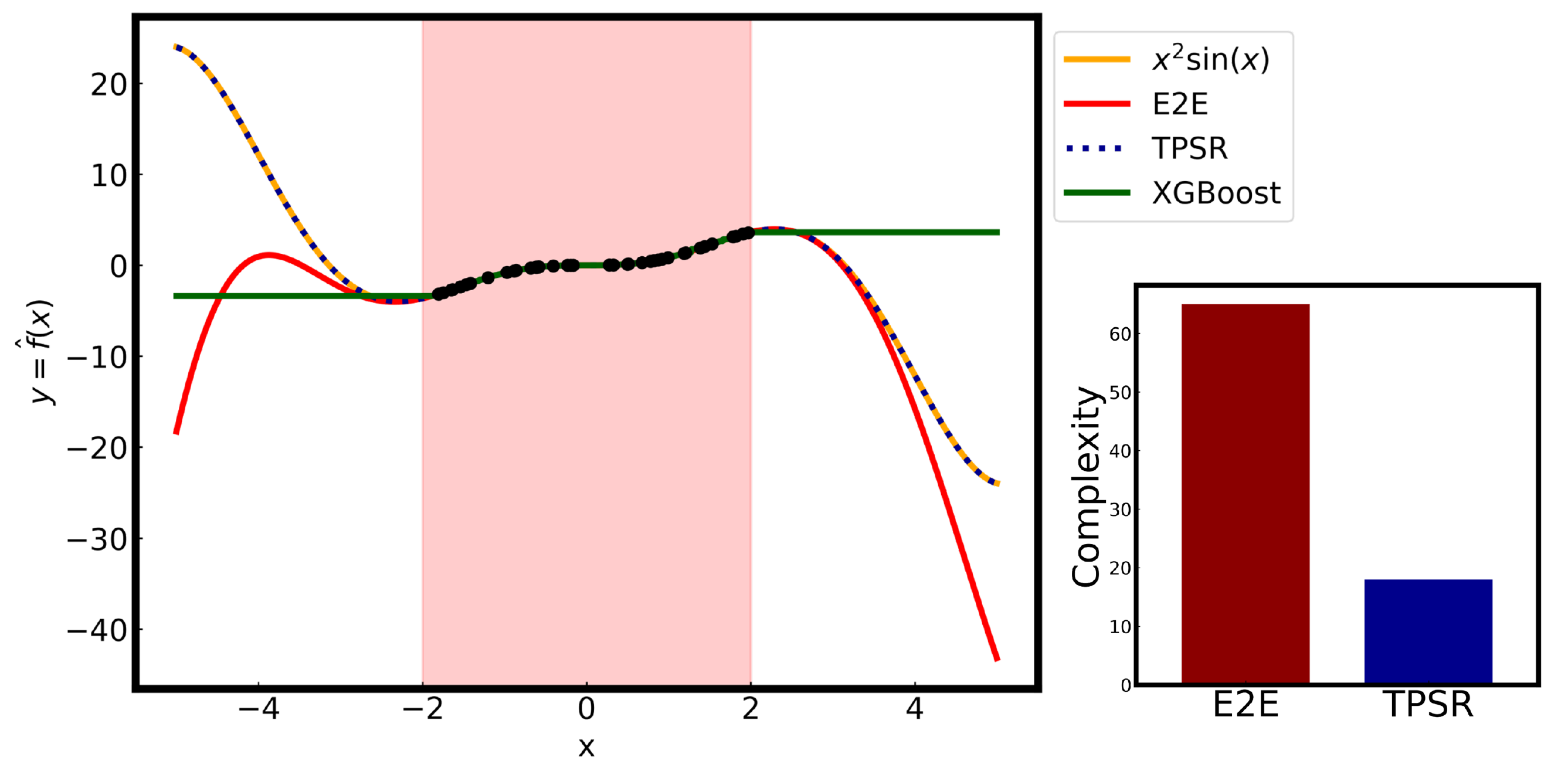}
\caption{Qualitative comparison of \modelname with E2E as well as black-box XGBoost \cite{XGBoost} fitting. The training dataset contains 200 points in range of $(-2,2)$ (shaded region), and the performance is evaluated over $(-5,5)$. E2E uses sampling decoding. }
\label{fig:qual}
\end{figure}

\vspace{-0.5em}
\subsection{Qualitative Study}
\label{sec:app-res-qual}
\vspace{-0.5em}
Fig.~\ref{fig:qual} offers a detailed qualitative analysis comparing the performance of \modelname, the E2E baseline (symbolic model), and XGBoost \cite{XGBoost} (black-box model) with respect to the ground-truth equation $x^2 \mathrm{sin}(x)$. The training dataset, depicted by the shaded red region, consists of 200 data points randomly sampled within the range of $(-2,2)$. The evaluation is performed on an out-of-domain region spanning from $(-5,5)$.

While all three models demonstrate a strong ability to fit the training data, the proposed \modelname method surpasses the E2E baseline in fitting the true underlying function, as evidenced by its performance in the out-of-domain region. This superior performance can be attributed to \modelname's capacity to generate less complex equations that still effectively fit the data, a feature highlighted in the accompanying complexity barplot. Moreover, the results showcase the general superiority of symbolic regression methods over the black-box XGBoost machine learning method when fitting the underlying function within the unseen evaluation range. This observation emphasizes the potential benefits of adopting symbolic regression techniques, such as \modelname, in providing more accurate representations of the data's underlying patterns and behaviors.

\vspace{-0.5em}
\subsection{Evaluating the Model-Agnostic Capability}
\label{sec:app-tpsr_nesymres}
\vspace{-0.5em}
In order to underscore the model-agnostic capabilities of \modelname, we also conducted evaluation experiments to include the integration of \modelname with the "Neural Symbolic Regression that Scales" (NeSymReS) model by Biggio \textit{et al.} \cite{Biggio-NeSymReS-ICML-2021}, a pioneering work for large-scale pre-training in SR.

\textbf{Limitations and Adjustments.}
NeSymReS, while influential, presents some inherent limitations: (1)~\textit{Dimensionality Constraint:} It can only handle datasets having a maximum of three dimensions ($D\leq 3$). This limits its application in wider experimental scenarios. (2)~\textit{Skeleton Prediction:} NeSymReS is also trained to only predict equation skeletons. As such, the system requires a more complex constant optimization process, further complicating its integration.

\textbf{Experiment Setup.}
Due to the constraints highlighted above, to evaluate the combination of \modelname with NeSymReS, we use a dataset composed of $52$ Feynman equations, as in \cite{Biggio-NeSymReS-ICML-2021}, ensuring the dimensionality constraint ($D \leq 3$) is respected.

\textbf{Results.} 
As illustrated in Table~\ref{table-tpsr-nesymres}, integrating \modelname with NeSymReS resulted in marked improvement. Specifically, results show that \modelname has significantly improved the fitting accuracy of NeSymReS without changing the average complexity of the equations when 
$\lambda=0.1$ and with a slight increase when $\lambda=0$.

\begin{table}[!ht]
\centering
\caption{Fitting accuracy and complexity performance of NeSymReS \cite{Biggio-NeSymReS-ICML-2021} with and without the proposed \modelname planning on $52$ \textit{Feynman} datasets with $D\leq3$.}
\begin{tabular}{lcc}
\toprule
Model                    &  Avg. ($R^2 >0.99$) $\uparrow$                  &  Avg. Complexity   $\downarrow$   \\
\midrule
NeSymReS   & 0.635    & \textbf{9.98 }    \\
NeSymReS+TPSR ($\lambda$=0.1)  & 0.808  & \textbf{9.98}  \\
NeSymReS+TPSR ($\lambda$=0)  & \textbf{0.827}         & 13.30  \\
\bottomrule
\bottomrule
\end{tabular}
\label{table-tpsr-nesymres}
\end{table}

\vspace{-0.5em}
\subsection{Additional SRBench Results}
\label{sec:app_srbench}
\vspace{-0.5em}

\begin{figure}[!ht]
\centering
\includegraphics[width=\linewidth]{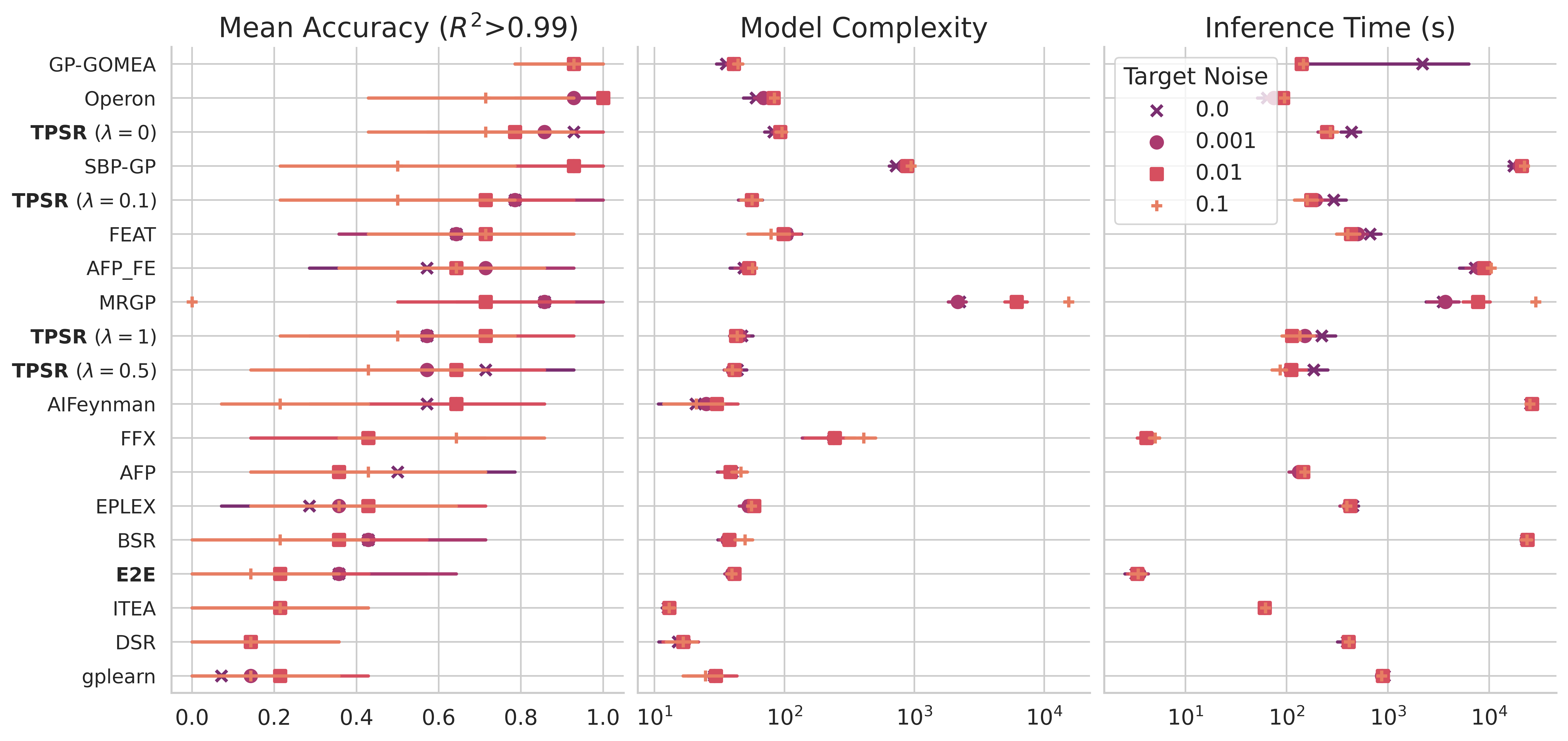}
\caption{Performance comparison of \modelname and SRBench algorithms in terms of Accuracy-Complexity-Time on \textit{Strogatz} dataset. Models are sorted based on mean accuracy defined as the ratio of solutions with $R^2>0.99$ on test set under various noise levels. The error bars represent the $95\%$ confidence interval.}
\label{fig:srbenchstrogatz}
\end{figure}

\paragraph{Strogatz Datasets. }
\label{sec:app_srbench_stro}
Figure \ref{fig:srbenchstrogatz} presents a performance comparison of \modelname and SRBench algorithms on the \textit{Strogatz} dataset (similar to the results shown for \textit{Feynman} and \textit{Black-box} datasets in Fig. \ref{fig:feynman_parigrid}). The \textit{Strogatz} dataset comprises 14 equations from a two-state system following a first-order ordinary differential equation (ODE). As it can be observed, E2E performance is less well on this dataset compared to other GP-based models due to the unique time-ordered distribution of observations, which differs substantially from the E2E's pre-training data. Notably, despite not being exposed to time-ordered data during pre-training, \modelname with the E2E pre-training backbone significantly enhances its performance on the Strogatz dataset. \modelname ranks among the top three baselines for fitting performance while maintaining comparable or even slightly superior levels of equation complexity and inference time.

\begin{figure}[!ht]
\centering
\includegraphics[width=\linewidth]{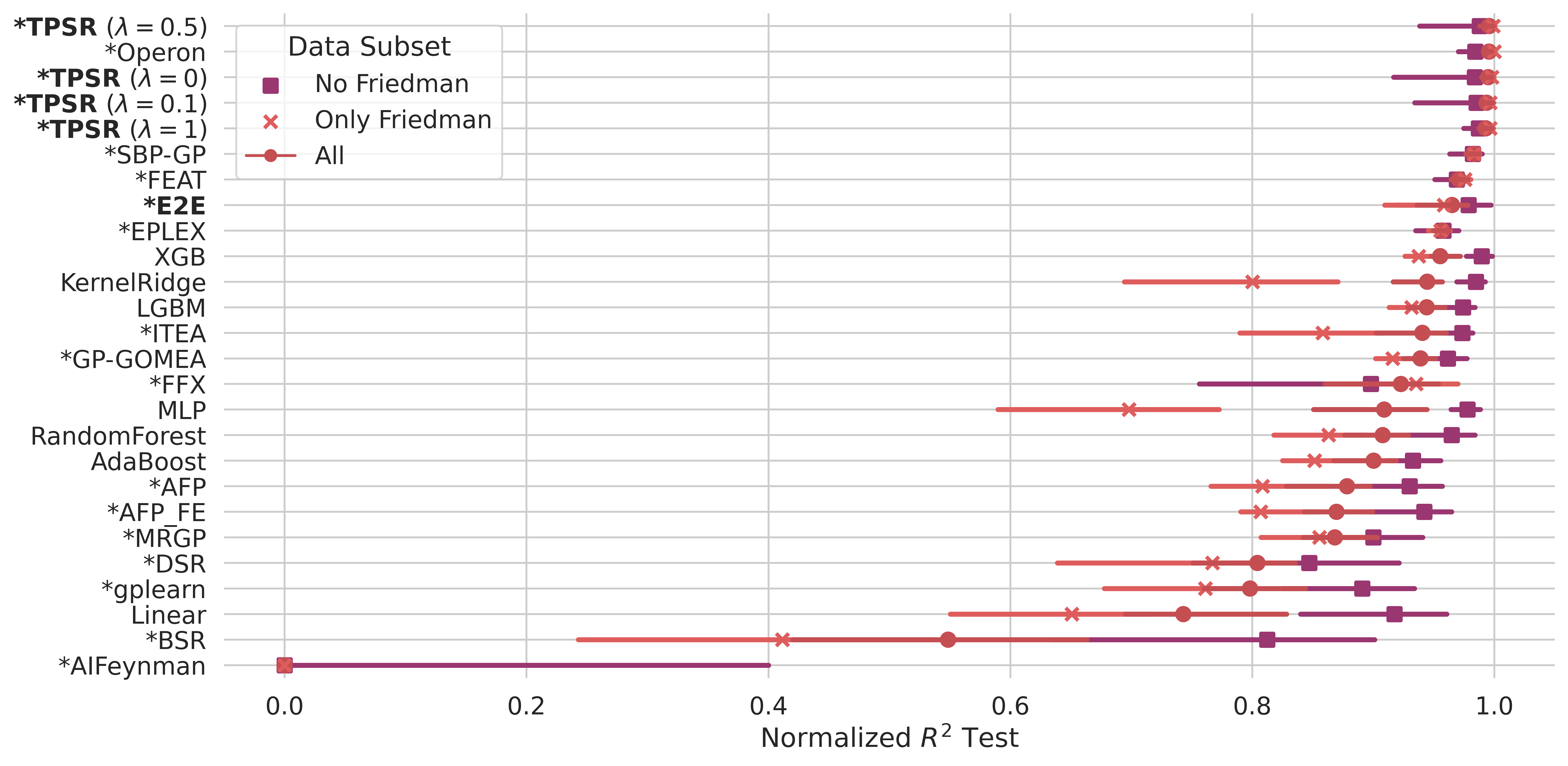}
\caption{Detailed performance comparison of \modelname and SRBench algorithms in terms of Accuracy (Fitting Performance) on \textit{Black-box} dataset groups: (1) Friedman \cite{friedman2000greedy} synthetic datasets, (2) non-Friedman datasets, and (3) all the black-box datasets. The error bars represent $95\%$ confidence interval and $"*"$ refers to SR methods vs. other ML methods.}
\label{fig:srbenchfriedman}
\end{figure}

\paragraph{Black-box Datasets.}
\label{sec:app_srbench_bb}
SRBench \cite{SRBench-Cava-NeurIPS-2021} studied black-box problems, originally extracted from OpenML \footnote{\url{https://www.openml.org/}} and integrated into PMLB \cite{Olson2017PMLB} datasets, include several datasets derived from Friedman's \cite{friedman2000greedy} synthetic benchmarks. 
These Friedman datasets, generated through non-linear functions, display varying degrees of noise, variable interactions, and non-linearity. 
As observed in earlier studies \cite{SRBench-Cava-NeurIPS-2021}, the results from the Friedman datasets tend to highlight the performance differences among top-ranked methods more noticeably than other benchmarks, where top-performing methods often deliver similar results.
Fig. \ref{fig:srbenchfriedman} shows that performance of several baselines such as KernelRidge, MLP, DSR, BSR, gplearn, and AFP, degrades on Friedman datasets. However, our \modelname variants maintains its superior performance across these challenging Friedman synthetic datasets and the remaining PMLB black-box datasets, asserting its state-of-the-art (top-$1$) status. Following \cite{SRBench-Cava-NeurIPS-2021}, this performance distinction is illustrated in Fig. \ref{fig:srbenchfriedman} with more details, separating the results of Friedman datasets from the rest of PMLB datasets.

\begin{figure}[!ht]
\centering
\includegraphics[width=\linewidth]{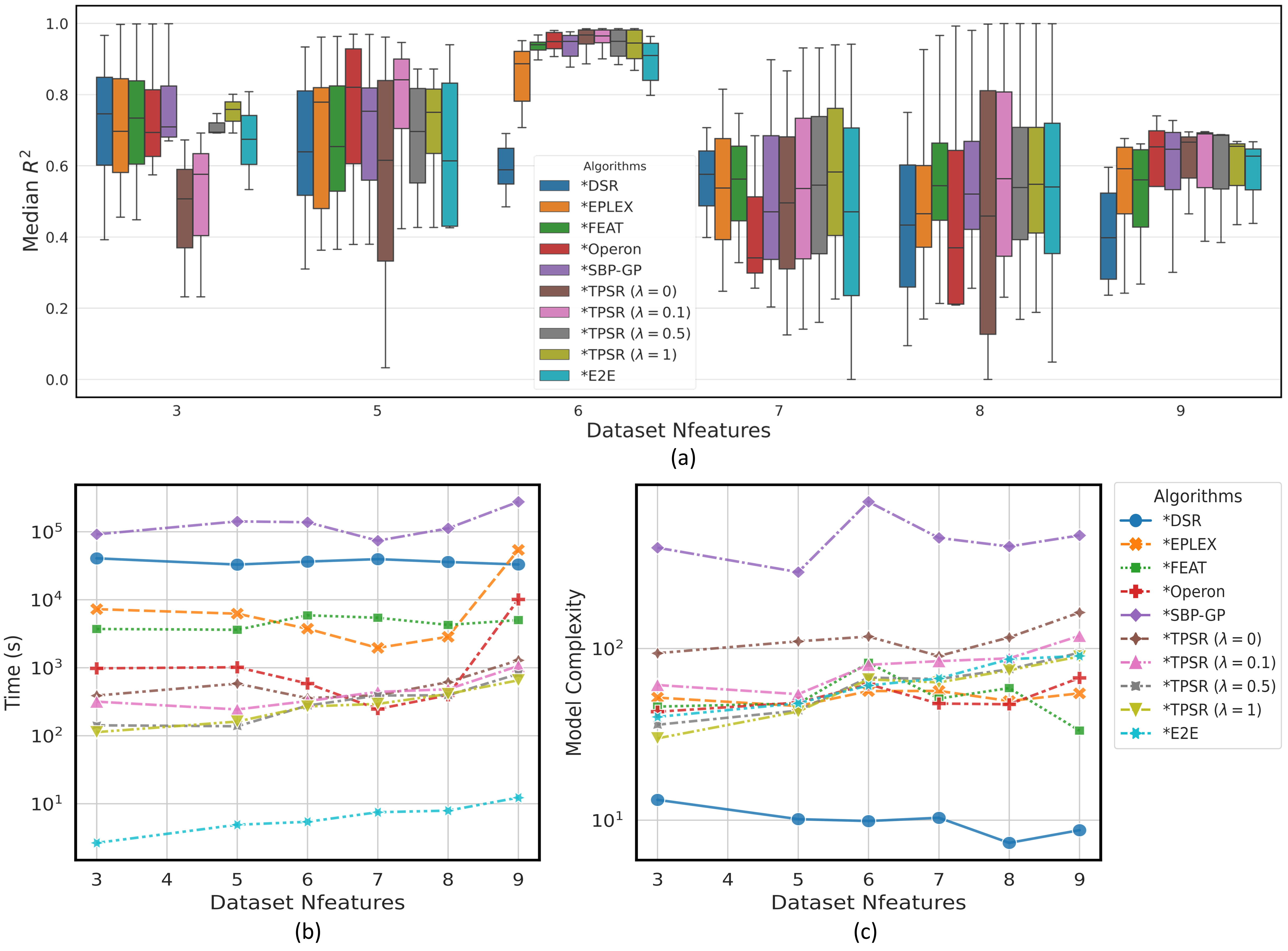}
\caption{Detailed performance comparsion of \modelname and competing baselines in terms of Accuracy-Complexity-Time metrics for \textit{Black-box} datasets of varying input dimensions.}
\label{fig:srbenchblackdim}
\end{figure}

Fig. \ref{fig:srbenchblackdim} shows an in-depth comparison of \modelname performance, varying $\lambda \in {0,0.1,0.5,1}$, against top competitors (Operon, SBP-GP, FEAT, EPLEX, and E2E) on \textit{Black-box} datasets of different input dimensions. Given E2E's pre-training on $d_{max}\leq10$, we focused on datasets with input dimensions below 10.
In Fig. \ref{fig:srbenchblackdim}(a), we note that dataset distribution and model performance both depend on the input dimensionality. \modelname consistently outperforms competitors across most dimensions. Interestingly, lower dimensions (e.g., $d=3$) favor \modelname with higher $\lambda=0.5,1$, resulting in better performance, while larger dimensions (i.e., $d=8,9$) benefit from smaller $\lambda=0,0.1$. This pattern aligns with the expectation that greater $\lambda$ values yield less complex expressions, more prevalent in lower dimensions, and vice versa.
Fig. \ref{fig:srbenchblackdim}(b) presents the average inference time for each model across different input dimensions. E2E is the fastest, while SBP-GP and DSR are the slowest. Notably, as input dimension increases, the inference time of Operon and EPLEX significantly escalates, hitting the scale of $10^4$ and $10^5$ seconds respectively, while \modelname's time remains relatively constant, peaking at $10^3$ seconds or roughly 30 minutes for $d=9$, compared to Operon's 3 hours and SBP-GP's 30 hours. 
This shows how efficient \modelname is compared to GP methods in finding state-of-the-art best-fitting expressions. 
Finally, Fig. \ref{fig:srbenchblackdim}(c) shows the average complexity of expressions generated by each model for different input dimensions. DSR's expressions are the least complex, while SBP-GP's are the most. \modelname with $\lambda=0$ is slightly more complex than its counterparts. Interestingly, \modelname with $\lambda=0.5,1$ produces less complex expressions than GP-based models like Operon, FEAT, and EPLEX at lower dimensions. However, as dimensions increase, these models generate less complex expressions than those of \modelname.

\vspace{-0.5em}
\subsection{Additional In-Domain Results}
\begin{figure}[!ht]
\centering
\includegraphics[width=\linewidth]{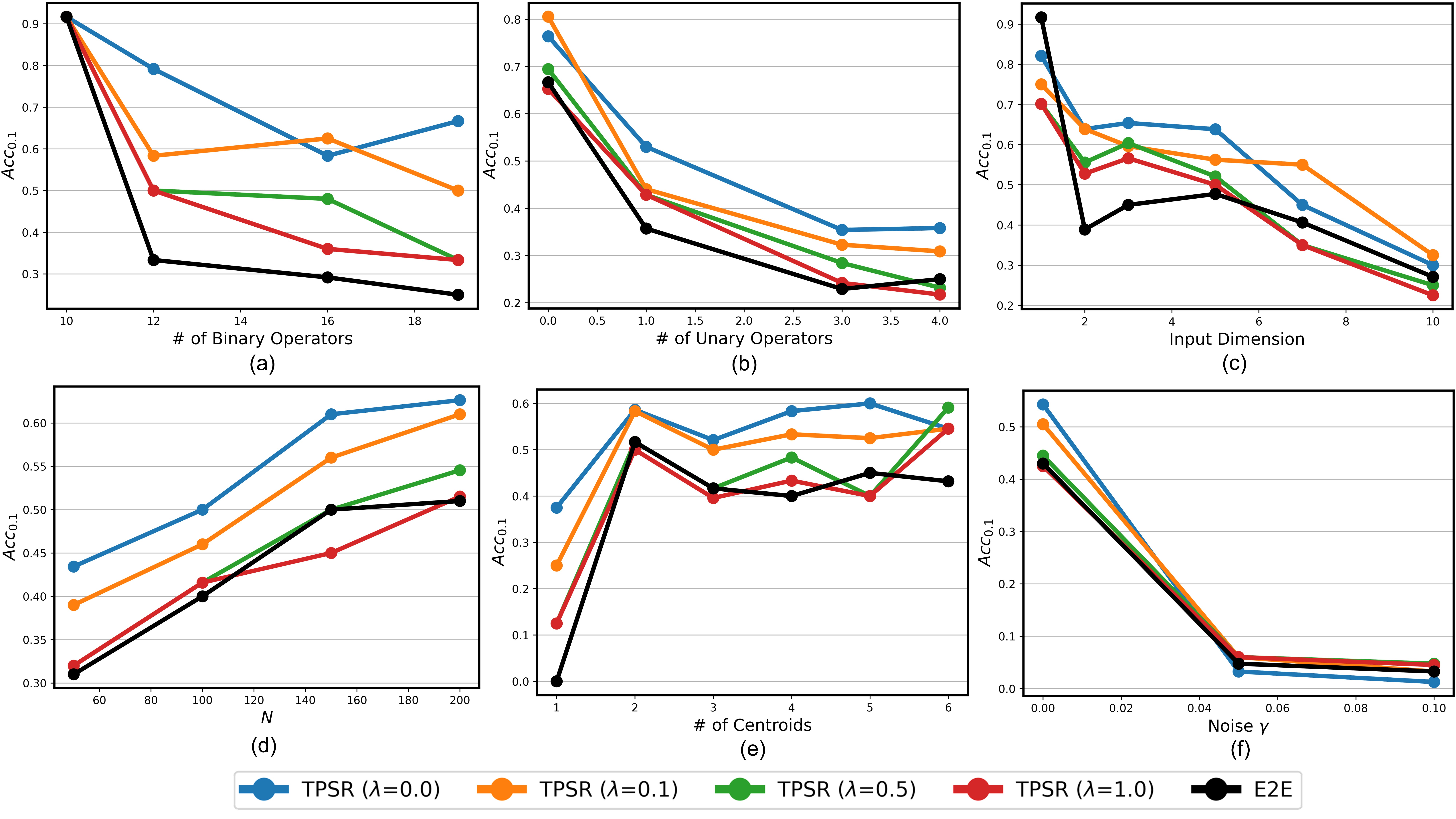}
\caption{Performance comparison of TPSR for varying $\lambda \in \{0, 0.1, 0.5, 1\}$ and E2E with sampling across different levels of formula and input difficulties: \textbf{(a) number of binary operators, (b) number of unary operators, (c) input dimension, (d) number of input points $N$ (e) number of input centroids, and (f) input noise variance $\gamma$.} }
\label{fig:indomainextra}
\end{figure}

\vspace{-0.5em}
Fig.~\ref{fig:indomainextra} presents a comprehensive performance comparison between our proposed TPSR method with varying controllable parameter $\lambda \in {0, 0.1, 0.5, 1}$ and the E2E baseline employing sampling for the \textit{In-domain Synthetic Dataset}.
As observed, when the complexity of the synthetic formula increases (as shown in the top row), such as increasing the number of binary/unary operators or the input dimension, the performance across all models tends to degrade.
However, we can see that TPSR with $\lambda=0,0.1$ ``always`` have lower performance drops and TPSR with $\lambda=0.5,1$ ``mostly`` have lower performance drop than the E2E. 
This highlights that not only does the incorporation of performance feedback in TPSR's MCTS-guided decoding help the transformer generation scale better with these difficulty levels, but the controllable complexity parameter $\lambda$ also plays a pivotal role in performance scaling for more challenging input functions.

Fig.~\ref{fig:indomainextra}(d) illustrates that the performance of all models increases as the number of input data points $N$ grows, as one would expect. However, TPSR with $\lambda=0,0.1$ exhibits considerably better low-resource performance for $N<100$ compared to the E2E model. It is important to note that the maximum $N_{max}=200$ since the E2E model is pretrained with $N\leq200$, and the transformer architecture employed in the encoding stage demands significant computational and GPU resources for training the model with $N>200$.

Fig.~\ref{fig:indomainextra}(e) also reveals that the performance of all models improves as the number of input data centroids increases, meaning that as the input data is sampled with greater diversity across different distribution clusters. We can clearly observe that our proposed TPSR with $\lambda=0,0.1$ consistently outperforms the E2E model, both with smaller and larger numbers of centroids.

Fig.~\ref{fig:indomainextra}(f) further investigates the impact of introducing multiplicative noise with variance $\gamma$ to the target $y$: $y \rightarrow y(1+\sigma), \sigma \sim \mathcal{N}(0,\gamma)$. As evident from the figure, the performance of all models deteriorates as the noise variance increases. This phenomenon highlights the sensitivity of the pretrained models to the input noise of the target variable. However, it is noteworthy that TPSR with $\lambda>0$ demonstrates slightly better performance compared to the E2E model, particularly when encountering larger noise variances.

\vspace{-0.5em}
\subsection{Additional Ablation Studies}
\label{sec:app-beta-ablation}
\begin{figure}[!ht]
\includegraphics[width=\linewidth]{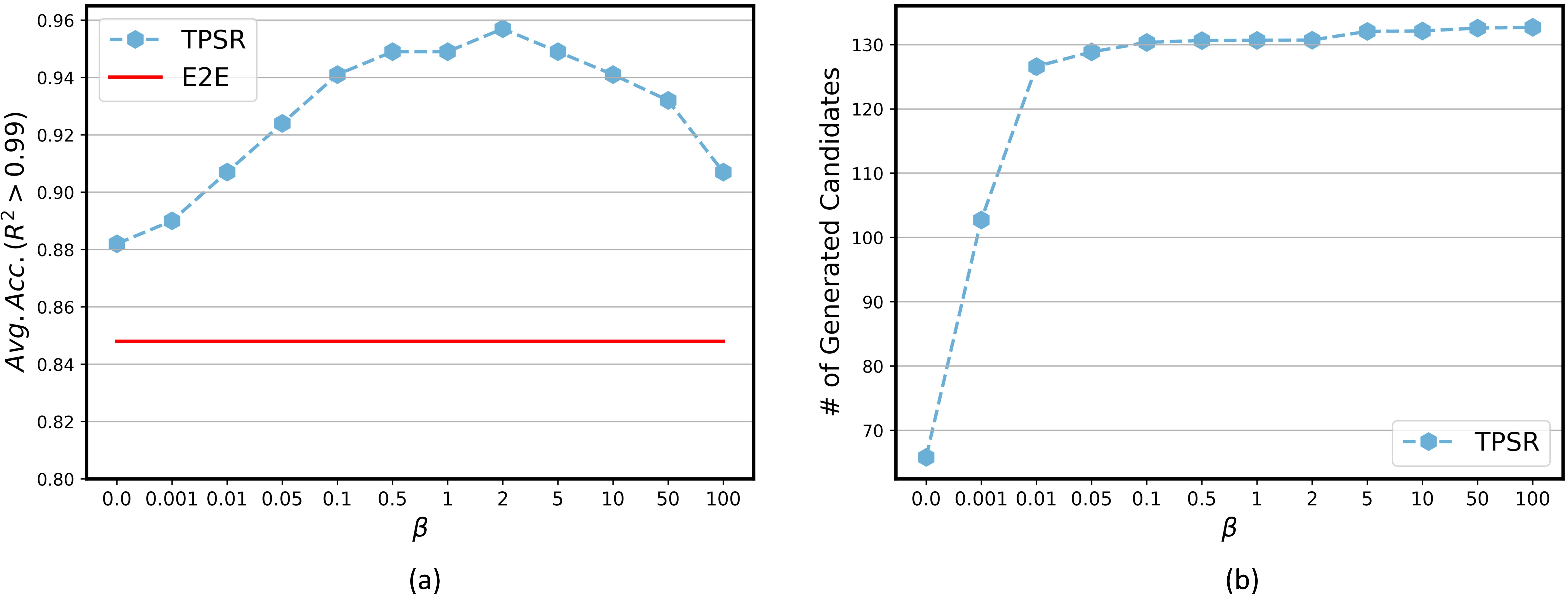}
\caption{Ablation study of $\beta$ on TPSR using $119$ \textit{Feynman} datasets: Balancing Exploration and Exploitation.}
\label{fig:ablationbeta}
\end{figure}

\vspace{-0.5em}
The selection of $\beta$, in Eq.~\ref{eq:uct} can also affect the exploration-exploitation trade-off, influencing the overall performance of \modelname. Fig.~\ref{fig:ablationbeta} demonstrates the impact of varying $\beta$ on TPSR's performance over $119$ \textit{Feynman} datasets, emphasizing the balance between exploration and exploitation. Based on the results, we observe that for small values of $\beta$, specifically $\beta=0$, the performance is sub-optimal. This diminished performance can be attributed to constrained exploration. Without sufficient exploration, the model might miss potential solutions or equation sequences that might be more effective. At the other end of the spectrum, with large values like $\beta=100$, there is also a decline in performance. This degradation can be linked to an over-emphasis on exploration at the cost of exploitation. By exploring too much without adequately leveraging the learned knowledge, the model can get overwhelmed with possibilities, some of which might not be beneficial. Experiment results highlight that optimal performance is achieved for $\beta$ values ranging between 0.1 and 10. As seen in Fig.~\ref{fig:ablationbeta}(b), with an increase in $\beta$, the number of equation sequence candidates grows, indicating an increase in exploration. However, beyond $\beta>0.1$, the increase in sequence candidates is marginal. This plateau suggests the possible activation of caching mechanisms due to repetitive sequence generation. Fig.~\ref{fig:ablationbeta}(a) also shows fitting performance against different $\beta$ values, illustrating the aforementioned trends and offering a visual guide for selecting $\beta$.

\begin{table}[!ht]
\centering
\caption{\label{table-SymExample} Example comparisons of symbolic expressions generated by E2E and \modelname, along with their respective fitting performance. 
}
\renewcommand{\arraystretch}{1.4}
\resizebox{0.7\columnwidth}{!}{
\begin{tabular}{ccc}
\toprule
&Expression& $R^2$ \\
\midrule
True Function & $2 X_0 (1-\cos{(\bm{X_1 X_2})})$ & \textendash \\
E2E Generation & $ CX_0 \left(C+C\cos{\left(CX_2+CX_1\right)}\right)+C$ & 0.453\\
TPSR Generation & $CX_0 \left(C+C\cos{\left(CX_1+CX_2\bm{+C X_1 X_2}\right)}\right)+C$ & \textbf{1.0}\\
\midrule
True Function & $\bm{\sin^2}{\left(\frac{\bm{X_0 X_1}}{\left(\frac{\bm{X_2}}{2\pi}\right)}\right)}$ & \textendash \\
E2E Generation & $ C\sin{\left(CX_0+CX_1+CX_2\right)}+C$ & 0.178\\
TPSR Generation & $C\bm{\sin^2}{\left(\frac{\bm{CX_1 X_0}}{\bm{CX_2} + CX_1}\right)}+C$ & \textbf{0.671}\\
\midrule
True Function & $X_0\left(\cos{\left(\bm{X_1 X_2}\right)}+\bm{X_3}\bm{\cos^2{\left(X_1 X_2\right)}}\right)$ & \textendash \\
E2E Generation & $CX_0\left(CX_3+CX_2+CX_1+C\cos{\left(CX_2+CX_1\right)}\right)^2 +C
$ & 0.878\\
TPSR Generation & $CX_0\left( \cos{\left(\bm{CX_2 X_1}\right)} +X_3^2\cos^2{\left(\bm{CX_2 X_1}\right)} \right)+C$ & \textbf{0.996}\\
\midrule
True Function & $X_0 \frac{\bm{\sin^2}{(\frac{\bm{X_1 X_2}}{2})}}{\bm{\sin^2}{(\frac{\bm{X_2}}{2})}} $& \textendash \\
E2E Generation & $CX_0 + CX_0\left(C\sin{\left(CX_1+CX_2\right)} + CX^2_1\right)^2+C$ & 0.655 \\
TPSR Generation & $ CX^2_0 \bm{\left( \frac{\sin{\left(CX_2 X_1\right)}}{\sin{\left(CX_2\right)}} \right)^2}+C$ & \textbf{0.991}\\
\midrule
True Function & $\sqrt{\left(X^2_0 + X^2_1 -2X_0 X_1 \cos{\left(\bm{X_2-X_3}\right)}\right)}$& \textendash \\
E2E Generation & $\sqrt{\left(CX_0 + CX_1\right)^2 \cos{\left(CX_2 + CX^2_3\right)}}+C 
$ & 0.939 \\
TPSR Generation & $ \sqrt{\left( CX_1 X_0  
\cos{\left(\bm{CX_2 - CX3}\right)} - CX^2_1 X^2_0\right)} + C$ & \textbf{0.986}\\
\arrayrulecolor{black}\bottomrule
\arrayrulecolor{black}\bottomrule
\end{tabular}
}
\end{table}

\vspace{-0.5em}
\subsection{Examples of Generated Symbolic Expressions}
\label{sec:app_example}
\vspace{-0.5em}
Table \ref{table-SymExample} presents example comparisons of symbolic expressions generated by E2E using sampling and our proposed TPSR model with $\lambda=1$ for 200 observation points of given true functions.
To improve readability and simplify notation, all constants in the generated expressions are denoted with the token ``C``. The table highlights how TPSR-generated symbolic expressions are more closely aligned with the true functions than those generated by E2E. The aligned components are bolded in the table entries.  Additionally, the fitting performance $R^2$ of TPSR-generated equations is notably superior to that of E2E-generated expressions. This comparison demonstrates how TPSR's integration of feedback in transformer decoding can yield quantitatively and qualitatively improved expressions using the same model weights. Improved learning of expressions behind the data can enhance the interpretability of black-box prediction models, contributing to their extrapolation and generalizability.

\vspace{-0.5em}
\section{Discussion and Future Work}
\vspace{-0.5em}
\paragraph{Limitations.} 
While our methodology exhibits substantial potential, it is not without limitations. One limitation of our approach is the increased inference time of the \modelname in comparison to simpler decoding methods like beam search and sampling. This extended inference time is primarily due to the process of searching and incorporating performance feedback during the generation phase in \modelname's decoding process. Nevertheless, by exploiting the semantic knowledge of large-scale pre-trained SR model, \modelname's inference time remains lower than the majority of GP-based SRBench baselines. Another factor influencing \modelname's performance is the dependency on the learned priors of the pre-trained SR model. \modelname is also subject to the inherent structural limitations of the pre-trained SR model, such as constraints on input dimensionality, expression length, and vocabulary definition. For example, the E2E model is pre-trained with a maximum input dimension ($d_{max}$) of 10, which in turn limits the \modelname with the E2E backbone to $d \leq 10$. However, it's important to note that \modelname is a model-agnostic framework, implying potential integration with more advanced pre-trained SR models in the future.

\paragraph{Future Directions.}
An intriguing dimension in the symbolic regression revolves around out-of-distribution data. Pre-trained Transformer SR methods, distinct from their search-focused counterparts, train on vast synthetic equation datasets stemming from certain distributions. Essentially, this distribution is shaped by specific equation generators and sampling techniques. Hence, any data or equation not stemming from these generators could be viewed as out-of-distribution. Our experimentation evaluated \modelname and the pre-trained E2E model \cite{Kamienny-E2E-symbolic-NIPS-2022} across both in-domain and out-of-distribution datasets, as in the SRBench. A crucial observation was that \modelname, with lookahead planning, considerably elevates the pre-trained model's performance on out-of-distribution datasets, a trend most pronounced in SRBench comparisons (as illustrated in Table \ref{table-PMLB}). 
While pre-trained models offer the strength of utilizing prior knowledge from large-scale datasets, they can be limited when faced with data far removed from their training distribution or unique equation forms they have not encountered during training. \modelname offers a partial solution through its decoding-stage search and planning, but it is still limited to the inherent constraints of the pre-trained SR model's priors. Addressing this challenge is an intriguing avenue for future research. Possible strategies might involve fine-tuning SR model weights using non-differentiable rewards for the new out-of-distribution datasets.

\vspace{-0.5em}
\section{Broader Impacts}
\label{sec:app_impacts}
\vspace{-0.5em}
\paragraph{Potential positive impacts.} The proposed \modelname approach for symbolic regression using transformer-based models has significant implications for both the research and practical communities. By integrating Monte Carlo Tree Search (MCTS) into the transformer decoding process, \modelname enables the generation of equation sequences that balance fitting accuracy and complexity, addressing key challenges in symbolic regression. This has wide-ranging applications in science and engineering domains, where accurate and interpretable mathematical models are essential for understanding and predicting complex phenomena. The improved performance of \modelname over state-of-the-art methods enhances the usability and reliability of symbolic regression models, enabling researchers and practitioners to extract valuable insights from their data and make informed decisions.

Moreover, \modelname offers practical benefits by leveraging the efficiency of transformer-based models and the pretraining priors. The ability to optimize equation generation using \modelname enhances the efficiency and scalability of symbolic regression, making it more accessible in resource-constrained settings. This opens up opportunities for the adoption of symbolic regression in various domains, including scientific research, engineering design, and optimization problems. The impact of \modelname extends beyond symbolic regression, as the integration of MCTS and non-differentiable feedback into transformer-based models can inspire novel approaches in other fields where the combination of symbolic reasoning and machine learning is valuable. Overall, \modelname has the potential to advance the state-of-the-art in symbolic regression and contribute to scientific and technological advancements.

\paragraph{Ethical considerations.}
Symbolic regression makes it easier for anyone to understand underlying patterns behind the data and learn interpretable symbolic mathematical models for observations. This approach brings the potential for machine learning models to achieve a balance of high predictive performance and transparency, which is critically valuable in sectors such as healthcare, where the interpretability of models can directly influence life-saving decisions. However, as with any powerful tool, the ethical issues of its use must be considered carefully.
For example, while symbolic regression can yield life-saving insights in the hands of healthcare professionals, it can also be exploited for malicious purposes. It could be used to decipher patterns and relationships within data where privacy should be maintained, leading to potential breaches of confidentiality.  This becomes particularly concerning as symbolic regression techniques mature, enabling more effective comprehension of symbolic mathematical relationships behind data values.
To mitigate this risk, we need the development of a separate modules tasked with screening input data and denying requests where pattern extraction could lead to harmful outcomes. 

\end{document}